%% file: arxiv.tex
\documentclass{article}

\usepackage[preprint,nonatbib]{neurips_2026}
\usepackage[T1]{fontenc}
\usepackage[utf8]{inputenc}
\usepackage[round]{natbib}

\newif\ifarxiv
\arxivtrue
\input{preamble-shared}

\begin{document}

\title{FADO: Learning Causal Explanations with Prior-Fitted Networks}

\author{%
  Seth Flaxman\thanks{Correspondence to seth.flaxman@rptu.de} \\
  University of Oxford \\
  RPTU Kaiserslautern-Landau
  \And
  Jess Carr \\
  University of Oxford \\
  Ellison Institute of Technology
  \And
  Annabel Jakob \\
  University of Oxford
  \And
  Marek Masiak \\
  University of Oxford
  \And
  Dino Sejdinovic \\
  Adelaide University
}

\maketitle

\input{body}

\end{document}

%% file: preamble-shared.tex
\usepackage{amsmath,amssymb,amsthm}
\usepackage{comment}
\usepackage{booktabs}
\usepackage{colortbl}
\usepackage{graphicx}
\usepackage{xcolor}
\usepackage{tikz}

\definecolor{motifA}{HTML}{009E73}
\definecolor{motifB}{HTML}{CC79A7}
\definecolor{motifC}{HTML}{0072B2}
\definecolor{motifD}{HTML}{D55E00}
\definecolor{motifE}{HTML}{E69F00}
\newcommand{\motifnode}[2]{%
  \tikz[baseline=(mn.base)]{\node[circle,draw=black,line width=0.3pt,
    fill=#1,text=white,inner sep=0pt,minimum size=1.8ex]
    (mn){\resizebox{!}{0.85ex}{#2}};}}
\usepackage{hyperref}
\usepackage{cleveref}
\usepackage{multirow}
\usepackage{array}
\usepackage{tabularx}
\usepackage{placeins}

\newcommand{\Var}{\operatorname{Var}}
\newcommand{\E}{\mathbb{E}}
\newcommand{\R}{\mathbb{R}}
\newcommand{\dop}{\mathrm{do}}
\newcommand{\tauhat}{\hat{\tau}}

\newtheorem{proposition}{Proposition}

\hypersetup{
  colorlinks=true,
  linkcolor=blue!60!black,
  citecolor=green!50!black,
  urlcolor=blue!60!black
}

%% file: body.tex
\begin{abstract}

In regression and classification tasks with tabular datasets, most of the literature on explanation in ML (e.g. SHAP, permutation importance) has focused on predictive attribution: does feature $X_i$ help predict outcome $Y$? At its heart, predictive attribution captures correlation, not causation.  We instead ask a causal question: would manipulating feature $X_i$ change outcome $Y$?  We define a natural quantity to answer this question, the feature intervention effect: $\Delta_i = E[Y\mid do(X_i=+1)]-E[Y\mid do(X_i=-1)]$. In general, estimating this quantity from observational data alone would entail strong causal assumptions. We propose FADO (Feature-wise Amortized Do-Operators), an amortized inference method using prior-fitted networks trained on synthetic tabular datasets generated by structural causal models. During training, the synthetic data-generating process is known, so we can compute ground-truth intervention effects $\Delta_i$, and train a model to predict them at test time from a purely observational tabular dataset.  As we demonstrate through extensive experiments, our new causal explainability method is easy to train and provides better explanations under a range of settings for which standard predictive attribution fails, including proxy variables, confounders, colliders, and placebos.  We evaluate it on real, synthetic and semi-synthetic datasets, with and without hidden confounders, and compare it to existing predictive attribution methods and state-of-the-art causal inference methods.

\end{abstract}

\section{Introduction}
\label{sec:intro}

When an ML model makes a high-stakes prediction in finance or healthcare, it is
natural to ask ``why?'' and demand an explanation
\citep{goodman2017explanation}.  Indeed, the EU's new AI Act \citep{euaiact2024}
contains an article, effective August 2026, on the ``Right to Explanation of
Individual Decision-Making'' laying out a right to ``clear and meaningful
explanations'' from deployers of AI systems in high-risk settings.  The
extensive ML literature on explainability aims to answer the
question ``why did the model make this prediction?'' 

\begin{figure}[t]
  \centering
  \includegraphics[width=\linewidth]{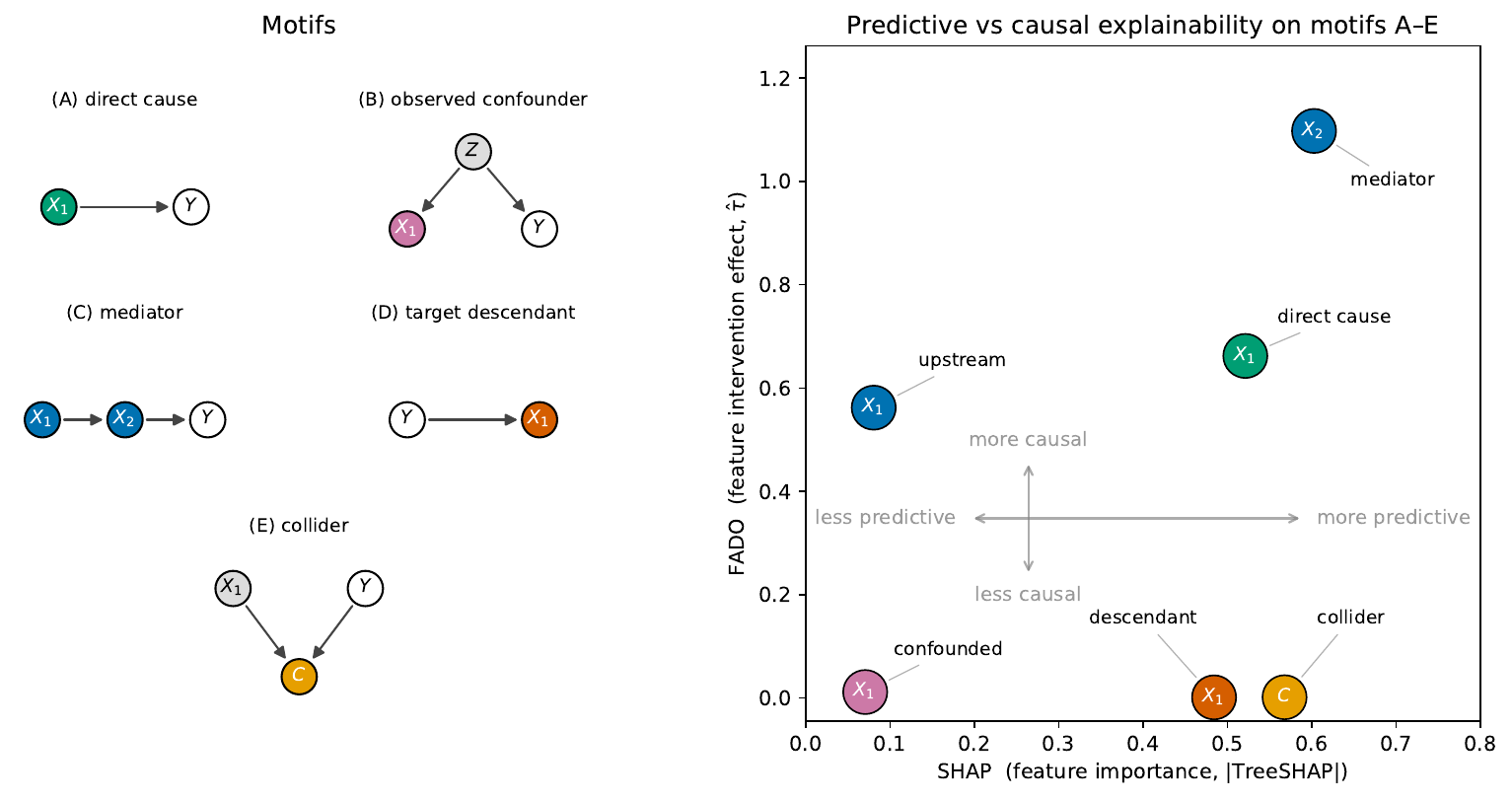}
  \caption[Five prototypical causal motifs where predictive and causal attribution disagree.]{Five prototypical causal motifs: small SCMs used to generate tabular data to
    demonstrate how predictive attribution based on association (SHAP) and causal
    attribution (our new method, FADO) can disagree. Direct cause \motifnode{motifA}{$X_1$} in motif A
    and mediator \motifnode{motifC}{$X_2$} in motif C are both causal and predictive.
    Target descendant \motifnode{motifD}{$X_1$} in motif D and collider \motifnode{motifE}{$C$} in motif E are predictive,
    but not causal. Confounded feature \motifnode{motifB}{$X_1$} in motif B is neither causal nor predictive.
  }
  \label{fig:pred-vs-causal}
\end{figure}

While these model-based predictive attribution questions are of course
incredibly important, machine learning is increasingly entering many scientific
domains, including natural science and social science. Here, a related but
distinct question arises: which variables would change the outcome if they were
intervened upon? By contrast, model-based predictive attribution asks: does a
given feature help predict the outcome? In scientific domains, it is often the
data-generating process from nature itself that is the object of study, rather
than a specific predictive model, so an explanation should answer ``what if''
questions about interventions \citep{pearl2009causality}, rather than
associational questions about a model's predictions. In practice, predictive
attribution methods give qualitatively different information from causal
inference methods (see \cref{fig:pred-vs-causal}) and we expect disagreements
between the two methods to be particularly useful for practitioners to interrogate,
as they could point towards non-causal associations driving model predictions.

Here, we focus on tabular datasets and view causal explanation through the lens of \textbf{feature intervention effects}: feature-wise interventional contrasts that ask how the outcome would change on average under an intervention on a particular predictor. Estimating this contrast for each feature yields a new causal feature-importance ranking, where importance reflects the effect of manipulating that feature in the data-generating process rather than its usefulness for predicting $Y$. It can be used alongside or
instead of popular predictive attribution methods like SHAP.

Existing explainability methods are usually
post-hoc: a model is fit and then explained. By contrast, we use prior-fitted
networks (PFNs) and amortization to learn to predict causal explanations
directly from underlying data-generating processes. This approach
is in line with a philosophy of ``learning to explain'' \citep{situ2021l2e}.
Building on a recent series
of papers on causal inference with PFNs
\citep{robertson2025dopfn,balazadeh2025causalpfn,ma2025causalfm,robertson2025fairpfn}
we demonstrate empirically that this approach is robust when 
trained on a sufficiently diverse set of priors. The training process is
very simple: we sample a tabular dataset from a data-generating process,
calculate the vector of feature intervention effects, which we term the {\bf
causal feature profile}, and then use supervised learning, training a transformer
to predict the causal feature profile from the tabular dataset.

\paragraph{Predictive vs.~causal attribution.}
\Cref{fig:pred-vs-causal} illustrates the difference between predictive and
causal attribution with five motifs, each a small structural causal model
(SCM). In motif A, $X_1$ is a direct cause of $Y$. In motif C, $X_2$ is a
mediator on the path from $X_1$ to $Y$. In the other three motifs, the
highlighted feature is associated with $Y$ but has no causal effect on
it: in motif B, $X_1$ co-varies with $Y$ only through their observed common
cause $Z$; in motif D, $X_1$ is caused by $Y$; and in motif E, the collider $C$
is caused by both $X_1$ and $Y$. The right panel plots each highlighted
feature's TreeSHAP importance under a LightGBM predictor
(\cref{sec:baselines}) against FADO's estimated intervention effect $\tauhat$.
We call a feature predictive if its TreeSHAP importance is high, and causal if
its true intervention effect is nonzero. The direct cause $X_1$ in motif A and
the mediator $X_2$ in motif C are both predictive and causal. The descendant
$X_1$ in motif D and the collider $C$ in motif E are predictive but not causal,
and FADO assigns them $\tauhat \approx 0$. The confounded $X_1$ in motif B is
neither predictive nor causal, since the predictor also sees $Z$. Finally, the
upstream $X_1$ in motif C is causal but not predictive, because its effect on
$Y$ is carried entirely by $X_2$, and FADO detects it. Predictive attribution
tells a user which features a model uses for predictions, while causal
attribution tells a user which features to intervene on. Manipulating $X_1$ in
motif D or $C$ in motif E would leave $Y$ unchanged. $X_1$ in motif C has low
TreeSHAP importance for predicting $Y$, but manipulating it would have a large
causal effect. \Cref{sec:baselines} compares predictive baselines with
ground-truth intervention effects on held-out SCMs, and
\cref{sec:shap-contrast} applies a standard TreeSHAP workflow to a real
randomized marketing dataset, where it highlights an outcome leak.

\paragraph{Contributions.}
\begin{itemize}
  \item We introduce the \textbf{feature intervention effect} $\tau_i$
        (\cref{eq:tau-main}) as a per-feature quantity
        for causal explanations, closely related to average
        treatment effects and dose-response curves
        (\cref{tab:target-vs-baselines}).
	It can be used to understand the impact of intervening on individual features,
	and to rank feature importance causally.

  \item We introduce a supervised learning framework called ``learning to explain'' and develop
	  \textbf{FADO} (\cref{sec:architecture}) for learning causal explanations, based on amortized inference
        over a large class of synthetic data generating processes
        (\cref{tab:scm-priors}), and train a foundation model
        (\cref{tab:checkpoints}) to predict a vector of feature
        intervention effects from a tabular dataset.
  \item We demonstrate that our method is effective, giving qualitatively
        different measures of variable importance versus existing predictive
        attribution methods (\cref{fig:pred-vs-causal,tab:motifs,sec:shap-contrast})
	and highly efficient (\cref{app:wallclock}), fast enough to run as an
	interactive demo in a web browser (\url{https://mlgh.net/fado-demo/})
	on tables of up to 64 features (\cref{app:wide-tables}).
  \item We theoretically analyze FADO, showing that squared-loss training
        leads to a Bayes optimal posterior mean (\cref{prop:bayes-action}).
\end{itemize}

FADO's aim is to output an accurate feature importance ranking based on causal interventions,
reliably identifying causal features and zeroing out non-causal features. While we compare to
causal discovery methods, our goal is not accurate point estimation of the causal 
effect for a specified treatment. Instead, we aim to simultaneously explain all features
at once.

\section{Setup and Target Functional}
\label{sec:method}

Let $X=(X_1, \ldots, X_p) \in\R^p$ and \(Y\in\R\) be the observed variables of
a structural causal model (SCM) \(M\), with possible additional latent
variables, and let \(G\) denote its directed acyclic graph
\citep{pearl2009causality,hernan2020causal}.
For each feature $i\in 1, \ldots, p$ we define the {\bf raw feature
    intervention effect} as the causal effect on $Y$ using do-operators:
\begin{equation}
  \boxed{
    \Delta_i(M)
    \;=\;
    \E_M[Y\mid\dop(X_i=b_i)]-\E_M[Y\mid\dop(X_i=a_i)]
  }
  \label{eq:delta-main}
\end{equation}
We define the {\bf standardized feature intervention effect} as:
\begin{equation}
  \boxed{
    \tau_i(M)
    \;=\;
    \frac{
      \Delta_i(M)
    }{
      \sigma_Y(M)
    },
  }
  \label{eq:tau-main}
\end{equation}

\begin{table*}[t]
  \centering
  \footnotesize
  \setlength{\tabcolsep}{4pt}
  \renewcommand{\arraystretch}{1.0}
  \caption{Existing estimators vs.~our proposed causal feature profile $\tau$.}
  \label{tab:target-vs-baselines}
  \begin{tabularx}{\linewidth}{@{}l l X@{}}
    \toprule
    Quantity                      & Definition                             & What it answers \\
    \midrule
    Marginal correlation          & $\mathrm{corr}(X_i,Y)$                 & Is $X_i$ associated with $Y$? \\
    Shapley value / SHAP          & $\phi_i$ averaged over subsets         & Does $X_i$ help a predictor of $Y$? \\
    Linear regression coefficient & $\hat{\beta}_i$ from $Y\sim X$         & Marginal linear contribution holding others fixed. \\
    Permutation importance        & $\Delta R^2$ when $X_i$ is permuted    & Predictive importance of $X_i$ in a fitted model. \\
    LiNGAM total effect           & $[(I-B)^{-1}]_{Y,i}$                   & Identified effect under linear non-Gaussian SCMs. \\
    Causal feature profile        & $\tau(M)$, see \cref{eq:tau-main} & Would intervening on $X_i$ change $Y$? \\
    \bottomrule
  \end{tabularx}
\end{table*}

where $\sigma_Y(M)$ is the standard deviation of $Y$ under $M$'s observational law.
For all features we use the standardized $(\pm 1)$ contrast
$(a_i,b_i)=(-1,+1)$\footnote{Input features are standardized
to mean 0 and standard deviation 1. Balanced binary features
are thus $\pm 1$, matching the convention of dividing inputs by $2\sigma$ before
reading off coefficients \citep{gelman2008scaling}.
In linear settings, the choice of $(a_i, b_i)$ is irrelevant, but
nonlinear mechanisms could be sensitive to this choice.
We conduct a sensitivity analysis in \cref{app:anchor-audit} and
find no meaningful change in the ranking of causal effects, and only
 small changes in the magnitude of the effects.}

  Our goal is to learn the entire $p$-dimensional
vector {\bf causal feature profile} $\tau(M)=(\tau_1(M),\ldots,\tau_p(M))$.
We write $\tauhat(D_n)$ for FADO's prediction of $\tau$ given an observational
table $D_n=\{(x_r,y_r)\}_{r=1}^n$, and $\tauhat_i$ for its $i$-th coordinate.

\input{architecture-section.tex}

\input{theory-section}

\section{Experiments}
\label{sec:experiments}
We evaluate FADO on an extensive battery of synthetic and real datasets,
and compare it with state-of-the-art causal inference and predictive
attribution methods (\cref{sec:baselines}). The evaluations cover held-out
SCMs and motif stress tests (\cref{sec:held-out-scms}), real covariates with
synthetic outcomes (\cref{sec:semisynth}), and real RCTs with constructed
proxy, leak and noise decoys (\cref{sec:real-rcts}). The
appendix adds  the Sachs DAG (\cref{app:sachs}), uncertainty analysis
(\cref{app:uncertainty}), architecture ablations (\cref{app:ablations}),
cross-family generalization (\cref{app:cross-family}) and timings
(\cref{app:wallclock}).

\subsection{Held-Out Random SCMs}
\label{sec:held-out-scms}

We evaluate FADO on held-out random SCMs from each of its eleven prior
families (\cref{tab:scm-priors}). Each family contributes $200$ SCMs with $p{=}5$
features padded to $15$, $n_{\mathrm{rows}}=512$ samples, and the per-column
noise of \cref{app:noise-families}. We report Pearson $r$ with the true
$\tau$, AUROC for nonzero-feature detection, and MAE on zero-effect features
(\cref{tab:held-out-scms}), pooled across the features of all $200$ SCMs,
with a marginal-correlation baseline ($\tauhat_i = 2\cdot\mathrm{Corr}(X_i,Y)$)
as a non-causal reference. The same FADO model is used in every row. Smaller
single-family variants, each trained and evaluated on one family, are
compared in \cref{app:held-out-extended}.

On five motifs (\cref{fig:pred-vs-causal}; A: direct cause,
B: observed confounder, C: mediator, D: target descendant,
E: collider), FADO
assigns $\tauhat \approx 0$ to non-causal predictive features (B/D/E)
while predictive baselines find non-zero effects;
per-motif paired-bootstrap CIs and per-baseline behaviour are in
\cref{tab:motifs} (\cref{app:motifs}).

Cross-family generalization (\cref{app:cross-family,fig:cross-family}),
seed stability (\cref{tab:seed-robust}), a heterogeneous-$p$ single
model on $p\in[3,15]$ (\cref{tab:heterogeneous-p}), and a scatter of
$\tauhat$ vs $\tau$ (\cref{fig:scatter-true-vs-pred}) are reported in
\cref{app:held-out-extended}.

\begin{table*}[t]
  \centering
  \caption{FADO on held-out random SCMs from each of its eleven prior
    families, $200$ SCMs per family, $p{=}5$
    padded to $15$, $n_{\mathrm{rows}}{=}512$, per-column noise of
    \cref{app:noise-families}. The same model is evaluated in every row.
    Family names follow \cref{tab:scm-priors}. The marginal baseline is
    $\tauhat_i=2\,\mathrm{Corr}(X_i,Y)$.}
  \label{tab:held-out-scms}
  \input{tables/foundation_per_family}
\end{table*}

\subsection{Predictive Attribution and Causal Inference Baselines}
\label{sec:baselines}

We compare FADO against three families of baselines on $500$ held-out
SCMs per regime, summarized in \cref{tab:all-baselines}.
\emph{Causal discovery} methods (DirectLiNGAM
\citep{shimizu2006lingam}, NOTEARS \citep{zheng2018notears}, PC
\citep{spirtes1991pc}, GES \citep{chickering2002ges}, and FCI
\citep{spirtes1995fci}) return a causal graph, from which $\tauhat$
is computed via $A=(I-B)^{-1}$ (\cref{app:linear-tau}). \emph{Per-column treatment effect}
estimators (Double Machine Learning \citep{chernozhukov2018dml} and
Causal Forest \citep{atheywager2019forest}, both doubly-robust by
construction) are designed for a known treatment, so we apply them
per-column with each feature playing the treatment role in turn.
\emph{Predictive attribution} baselines (marginal correlation, linear
regression, permutation importance, and path-dependent TreeSHAP via
LightGBM) answer a predictive rather than an interventional question. We include
them to quantify the difference between predictive and causal importance on a
causal target.
We also compare with Do-PFN and CausalPFN \citep{robertson2025dopfn,balazadeh2025causalpfn}. These methods require designating a single binary treatment column so the evaluation is not entirely fair: we binarize and use them one feature at a time (\cref{app:offlabel-pfn}).

FADO outperforms every other method on every column of
\cref{tab:all-baselines}, except where LiNGAM's identifiability
assumptions (linear non-Gaussian, no hidden confounders) hold.
We also compare FADO with do-Shapley \citep{jung2022measuring}, a
Shapley-value attribution of interventional effects that must be given the
true SCM. FADO, which sees only the observational table, beats it on three of
five families, while do-Shapley is better on the two hidden-confounder
families (\cref{tab:shapley-comparison} in \cref{app:held-out-extended}).

\begin{table*}[t]
  \centering
  \caption{FADO vs.~causal discovery, treatment effect, and
    predictive attribution baselines. Each entry is the Pearson $r$ between
    $\tauhat$ and $\tau$ across $500$ held-out SCMs. Bold marks the
    column winner when its paired-bootstrap lead over the runner-up is
    significant at $95\%$ (paired-$\Delta$ CIs in \cref{app:paired-delta}).
    $^\dagger$
    DML and Causal Forest were fit once per feature, with that feature
    playing the treatment role.}
  \label{tab:all-baselines}
  \resizebox{\linewidth}{!}{\input{tables/all_baselines}}
\end{table*}

To evaluate FADO and other methods' ability to zero out non causal features, we construct a diverse suite of real and semi-synthetic experiments, including RCTs, physical systems (\cref{app:chambers}), and synthetic decoys with real covariates.

\subsection{Semi-Synthetic Experiments: Real Covariates}
\label{sec:semisynth}
\label{app:acic-fp}

We test FADO on real covariates with simulated outcomes. We take the
covariates of the ACIC-2016 benchmark \citep{dorie2019acic}, $4802$ rows of
$58$ features from a real study, sample $8$ columns per dataset, and simulate
an outcome from them, so that every column's $\tau$ is known.
We created 40 SCM instances per outcome family: linear, mixed nonlinear,
and interaction.
We consider four types of decoys:
(1) a {\em proxy} variable is correlated with a
true causal predictor; (2) a {\em leak} variable is a function of the outcome variable;
(3) a {\em pure noise} variable is independent of the other variables;
and (4) a {\em confounder child} variable is a noisy copy of a hidden confounder.
In all cases, the true $\tau$ for these decoy variables is 0.
We considered two regimes: no confounding and hidden confounders.
We calculated false positives as
mean $|\hat\tau|$ at decoy positions; lower is better.

Our findings are in \cref{tab:acic-fp}. FADO assigns nearly zero effect to
pure-noise decoys (below $0.01$), as does multivariate regression. FADO assigns smaller effects than marginal
correlation to every decoy type, and it is the best method on the outcome leak
in every setting. Multivariate regression assigns the smallest effects to
proxies and confounder children, because conditioning on the true causes
absorbs them, but it gives the outcome leak a large effect.

\begin{table*}[t]
  \centering
  \caption{ Mean $|\hat\tau|$ at decoy positions on ACIC-2016 real covariates
    \citep{dorie2019acic} (true $\tau = 0$; lower is better), $40$ SCM
    instances per outcome family. F~=~FADO, Mg~=~marginal correlation,
    Mv~=~multivariate regression. Confounder child decoys exist only in the
    hidden-confounder regime.}
  \label{tab:acic-fp}
  \resizebox{\linewidth}{!}{\input{tables/acic_fp}}
\end{table*}

\subsection{Real RCTs with Constructed Decoys}
\label{sec:real-rcts}

\Cref{sec:semisynth} uses real covariates with simulated outcomes. Here we use real outcomes. The Hillstrom e-mail marketing experiment \citep{hillstrom2008} randomly assigned customers to receive a men's merchandise e-mail or no e-mail, and recorded whether they visited the website. We augment its pre-treatment covariates and treatment $T$ with the same kinds of constructed decoys as in \cref{sec:semisynth}: a proxy $P_T$ correlated with $T$, a leak $L_Y = Y + \varepsilon$, and three pure-noise columns. FADO sees the augmented covariates $\{X,T,P_T,L_Y,\mathrm{noise}\}$ and $Y$, with no indication of which column is the treatment. Success is defined as FADO estimating an effect for $T$ that matches the sign and magnitude of the RCT estimate, and estimating effects of approximately zero for $P_T$, $L_Y$ and the noise columns.

\label{sec:shap-contrast}
On the augmented Hillstrom feature matrix, we fit a LightGBM on
$\{X,T,P_T,L_Y,\mathrm{noise}\}$ and report path-dependent TreeSHAP via
LightGBM's \texttt{pred\_contrib} \citep{lundberg2017shap,lundberg2020tree} on visit
outcome\footnote{The choice of SHAP method is irrelevant in this case, because
\citet{janzing2020feature}'s ``interventional SHAP equals do-calculus'' result holds only under a flat causal graph in which every feature is a root node. See \cref{app:shap-variants}.}.
This conclusion is not specific to the SHAP variant used: all variants we test behave alike on our benchmarks (\cref{app:shap-variants}).
The classifier AUC is $0.998$,
driven entirely by the leak column $L_Y$, which SHAP scores as the
most important feature. FADO also assigns $L_Y$ a large effect
($\tauhat = +1.03$, $95\%$ CI $[+0.67,+1.32]$), as does DR-Learner
\citep{kennedy2020drlearner}, run per-feature with RandomForest
first/second/final stages and 2-fold cross-fitting
($\tauhat = +1.08$, $[+1.06,+1.09]$, $\sim 30\times$ its largest other
estimate). On the true (hidden) treatment FADO under-shoots
($-0.01$, $[-0.06,+0.09]$) consistent with the magnitude shrinkage
discussed in
\cref{sec:limitations}. Per-feature estimates and bootstrap CIs are in
\cref{tab:shap-contrast} (\cref{app:shap-contrast-table}).

We further test on $124$ real RCTs from RCT Bench \citep{shao2026covariates}, a public collection of participant-level trial data. We run FADO on each trial's covariates and treatment, without decoys, and compare its estimate for the treatment column, $\tauhat_T$, with the trial's difference in means. Though not designed for this task, FADO attains a correlation of $0.85$ with ground truth. CausalPFN, specifically designed for this task, achieves $0.96$.

\subsection{Uncertainty Quantification}
\label{sec:uncertainty}

\begin{figure}[t]
  \centering
  \includegraphics[width=.8\linewidth]{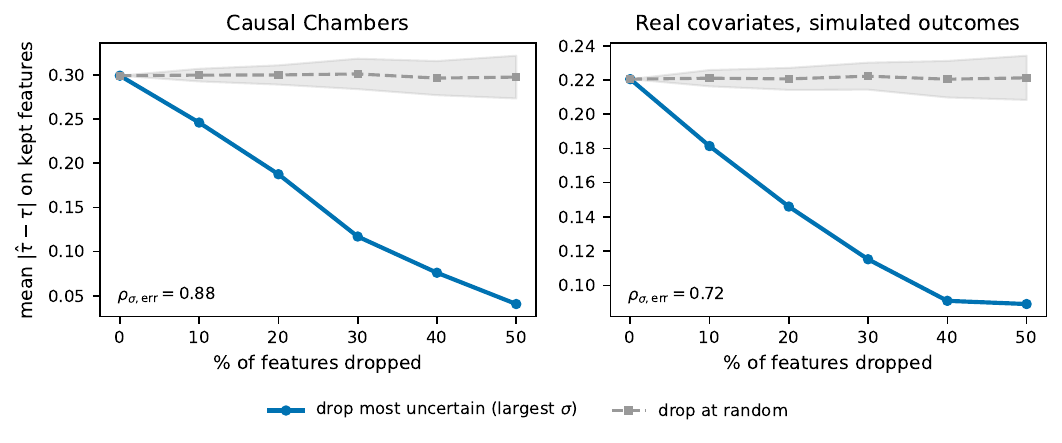}
  \caption{FADO's bootstrap uncertainty identifies its least accurate
    estimates. For each feature we compute the bootstrap standard deviation
    $\sigma$ of $\tauhat$, drop the given percentage of features with the
    largest $\sigma$ (blue) or the same number of features at random (grey;
    shaded: $\pm1$ sd over random draws), and report the mean error
    $|\tauhat-\tau|$ on the features kept. Error falls steadily when the most
    uncertain features are dropped and stays flat when features are dropped
    at random. $\rho_{\sigma,\mathrm{err}}$ is the Spearman correlation
    between $\sigma$ and the error. Left: the Causal Chambers light tunnel, a
    physical system with measured intervention effects
    (\cref{app:chambers}). Right: real covariates with simulated outcomes
    (\cref{sec:semisynth}).}
  \label{fig:selective-prediction}
\end{figure}

We quantify FADO's uncertainty with $100$ random bootstrap resamples, reporting feature-wise standard deviation $\sigma_i$ of $\tauhat_i$. Though confidence intervals are not calibrated (\cref{tab:uncertainty-coverage}), they do track error: Spearman correlation between $\sigma_i$ and $|\tauhat_i - \tau_i|$ ranges from $0.65$ to $0.90$ across regimes (\cref{tab:uncertainty-quality}), and leaving out the most uncertain features sharply reduces error, both on a physical system and on real covariates with simulated outcomes (\cref{fig:selective-prediction}).

\section{Related Work}
\label{sec:related}

Our work is situated in multiple literatures: explainability, causal
inference and causal discovery, amortized inference, and prior-fitted networks.
We mention foundational papers and those that are directly relevant to our
approach.  Permutation importance \citep{breiman2001random,fisher2019all} along
with SHAP \citep{lundberg2017shap,jethani2022fastshap} are widely adopted
predictive attribution approaches, focused on explaining the predictions of a
fitted model rather than the data-generating process. A method with an
overlapping spirit to ours is learning to explain (L2E) \citep{situ2021l2e}, which trains a network
to predict explanations directly, targeting a predictive rather than
interventional functional.

Causal inference is a long-standing area within statistics
(e.g.~\citep{holland1986statistics}), social science
(e.g.~\citep{morgan2014counterfactuals}), and epidemiology
(e.g.~\citep{hernan2020causal}), with a related and very rich literature on
causal discovery in computer science
(e.g.~\citep{pearl2009causality,spirtes2000causation,buhlmann2014cam,peters2017elements}).
Causal discovery takes a dataset and returns a causal graph. A separate, deep
literature on per-unit treatment effect estimation
\citep{chernozhukov2018dml,atheywager2019forest,kennedy2020drlearner,hill2011bayesian}
targets (conditional) average treatment effects for a user-designated
treatment, whereas FADO returns a length-$p$ vector of per-feature intervention
contrasts in a single forward pass, with no designated treatment.

Prior-fitted networks (PFNs) \citep{muller2022pfn} amortize Bayesian inference
by training a model on synthetic datasets sampled from a prior, so that the
trained network performs inference in-context at test time. For tabular data
this paradigm is typified by TabPFN \citep{hollmann2025tabpfn} and causal
inference using PFNs has seen much recent work
\citep{robertson2025dopfn,balazadeh2025causalpfn,ma2025causalfm,robertson2025fairpfn}.
Causal Articulation Theory \citep{zecevic2026cat} is complementary to our work, because it can produce natural language explanations from FADO's output.

\section{Limitations}
\label{sec:limitations}
FADO learns the Bayes posterior mean of $\tau(M)$ under the prior. Thus, its
accuracy in matching the true causal feature profile depends on whether that
profile is identifiable from tabular observational datasets
(\cref{sec:theory-identifiability}, A1--A3). For example, FADO's performance
suffers under linear Gaussian and hidden confounder data generating processes
(\cref{tab:all-baselines}) because observational datasets under this prior
are consistent with multiple causal structures.
We found that FADO provides a reliable feature ranking and identifies causal vs.\ non-causal features (including zeroing proxy/noise decoys), but it does not accurately predict point estimates of effect sizes.
While the paper's main FADO model is trained on tables of up to 15 features, we extend training to 64 features in \cref{app:wide-tables}.

\section{Conclusion}
\label{sec:conclusion}
We introduced a new method, FADO, motivated by the increasing use of machine
learning in scientific domains, where predictive attribution methods are useful
but can mislead on their own. We hope FADO becomes a default, baseline method
for causal explanations, responsive to ``what if?'' questions, phrased
interventionally. FADO's implementation requires amortized inference through
prior-fitted networks, building on the exciting ``learning to explain'' framework, 
which could be relevant for many application domains and types of data beyond tabular.

\subsection*{AI use statement}
We used Claude (various models) extensively during this work, to write and run
training, evaluation, and synthetic data generation code and baselines, to
generate tables and figures from experimental
results, to double check citations, and for \LaTeX{} editing and formatting.
We used ChatGPT (various models) to check proofs.
Final text was written by authors, partly based on AI-generated summaries of
experimental code and results. We take full responsibility for the correctness
of the paper's content.

\begin{ack}
This work was conducted as part of the EIT CDT in Fundamentals of AI, of
which Jess Carr is a student. Funding was provided by the Ellison Institute of
Technology. SF acknowledges support from the Oxford Martin School and Oxford
Global Health.
\end{ack}

\bibliographystyle{plainnat}
\bibliography{references}

\clearpage
\appendix
\thispagestyle{empty}

\onecolumn

\let\oldappendixsection\section
\renewcommand{\section}{\FloatBarrier\oldappendixsection}

\section{Code}
\label{app:code}

Code, trained checkpoints and the scripts that reproduce every table and figure are
available at \url{https://github.com/MLGlobalHealth/FADO}. An interactive browser
demo is at \url{https://mlgh.net/fado-demo/}.

\section{Held-out Random SCMs: Extended Results}
\label{app:held-out-extended}

\Cref{tab:held-out-variants} compares FADO with smaller single-family
variants (\cref{tab:checkpoints}), each trained on one prior family and
evaluated on $500$ held-out SCMs from that family. Every row is therefore an
in-distribution evaluation, and rows are not directly comparable. All rows
use Laplace noise in every column, except the Gaussian control. The FADO row
evaluates the FADO model on linear SCMs under this protocol. It is higher
than the linear row of \cref{tab:held-out-scms} ($0.99$ against $0.95$ in
Pearson $r$), which uses FADO's per-column noise of
\cref{app:noise-families}. \Cref{tab:held-out-dags-extended} adds further
variants: the Linear $p{=}5$ variant at $15$k steps, the two Large
architecture variants ($d{=}192$, 4+4 layers, part of the architecture sweep
in \cref{app:ablations}), the Gaussian control at $p{=}8$, MLP $+$ hidden
confounder, and the $p{=}13$ replicates of the hidden-confounder and
mixed-type families. All results in the paper use the FADO model except the
tables comparing model variants
(\cref{tab:held-out-variants,tab:held-out-dags-extended,tab:heterogeneous-p,tab:sachs-ckpts,tab:checkpoints}).

\begin{table}[h]
  \centering
  \caption{Single-family variants on held-out random SCMs, $500$ per row.
    Each variant from \cref{tab:checkpoints} is evaluated on its own
    training family, so rows are not directly comparable. The FADO row
    evaluates the FADO model on linear SCMs under the same protocol, with
    Laplace noise in every column.}
  \label{tab:held-out-variants}
  \input{tables/held_out_dags}
\end{table}

\begin{table}[h]
  \centering
  \caption{Further single-family variants on held-out random SCMs, $500$
    per row. Companion to \cref{tab:held-out-variants}, with the same
    evaluation protocol.}
  \label{tab:held-out-dags-extended}
  \input{tables/held_out_dags_extended}
\end{table}

\begin{figure}[h]
  \centering
  \includegraphics[width=\linewidth]{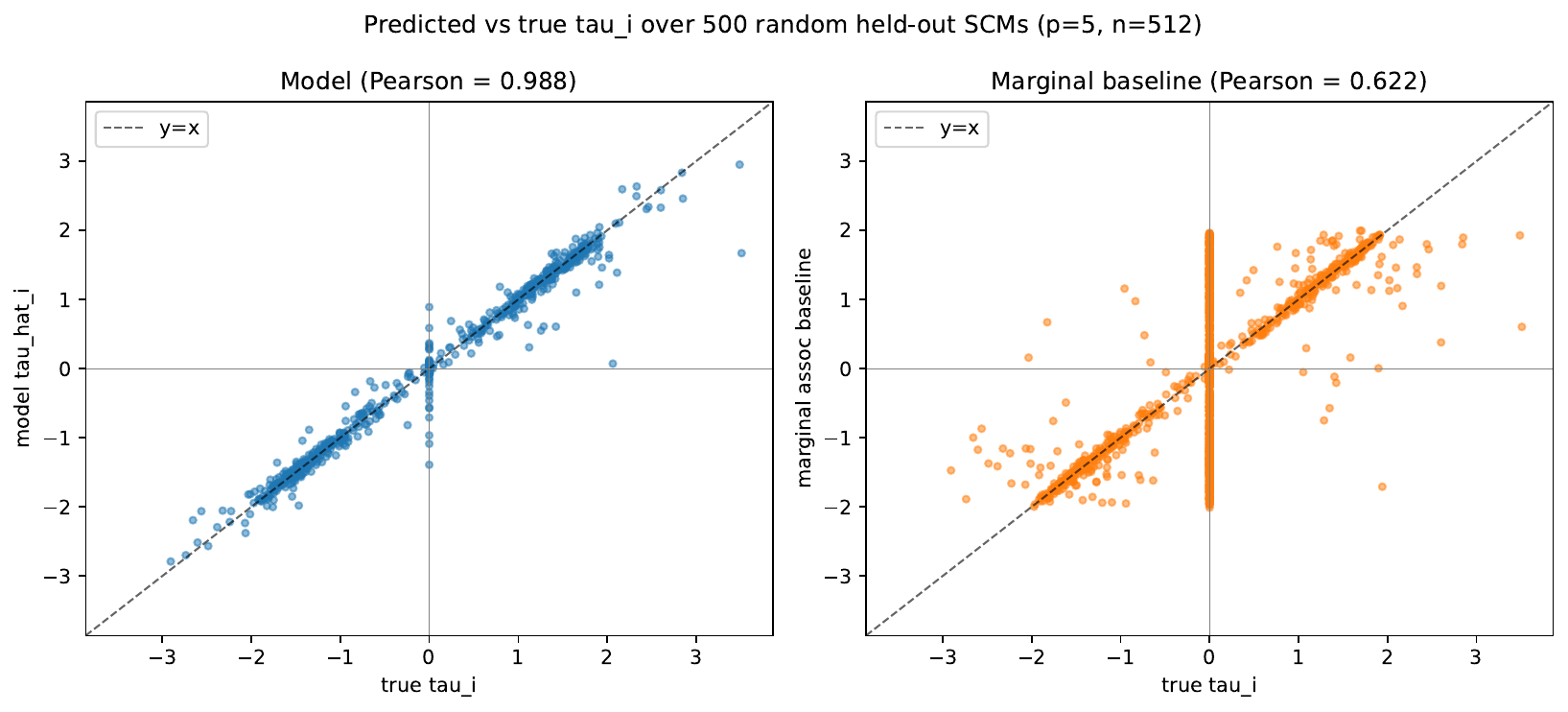}
  \caption{Predicted $\tauhat_i$ vs true $\tau_i$ across $500$ held-out
    random SCMs ($p{=}5$ linear non-Gaussian Laplace,
    $n_{\mathrm{rows}}{=}512$). Left: FADO $\tauhat$, Pearson $r$ reported
    in the panel title; right: marginal-correlation baseline. FADO
    tracks the $y{=}x$ diagonal; marginal does not.}
  \label{fig:scatter-true-vs-pred}
\end{figure}

\begin{table}[h]
  \centering
  \caption{Seed robustness: same config, three seeds.}
  \label{tab:seed-robust}
  \input{tables/seed_robust}
\end{table}

\begin{table}[h]
  \centering
  \caption{Heterogeneous-$p$ single model, $p\!\in\![3,15]$ linear SCMs.}
  \label{tab:heterogeneous-p}
  \input{tables/heterogeneous_p}
\end{table}
We compare do-Shapley, which has access to the true graph, to FADO, which does not (\cref{tab:shapley-comparison}).
\begin{table}[h]
  \centering
  \caption{FADO against Shapley methods given the true causal model, on
    held-out random SCMs ($p{=}5$; $200$ SCMs per family, $80$ for the Monte
    Carlo families). Cells are Pearson $r$ between each method's per-feature
    attribution and the true $\tau$. $\dagger$: do-Shapley
    \citep{jung2022measuring} and Causal Shapley \citep{heskes2020causal} are
    given the true SCM, both graph and mechanisms. DSCM is given the graph and
    learns the mechanisms from data. FADO sees only the observational data.
    Marginal and Multivariate are the associational baselines of
    \cref{tab:all-baselines}.}
  \label{tab:shapley-comparison}
  \input{tables/shapley_comparison}
\end{table}

\section{Wider Tables}
\label{app:wide-tables}

The FADO foundation model of the main text is trained on tables with
$p\in[3,15]$ features, padded to $15$ columns. Nothing in the architecture
depends on $p$ (\cref{sec:arch}), so the same network can be trained on
wider tables. We trained two further models with the same architecture,
optimizer, step count and prior families as \cref{tab:checkpoints}, differing
only in how table width enters training:

\begin{itemize}
  \item \textbf{FADO $p{\le}64$.} Following the width scheme of
        TabPFN~\citep{hollmann2025tabpfn}, each training step draws a single
        width $p\sim\mathrm{LogUniform}[3,64]$ and samples a batch of SCMs at
        that width, with no padding. The number of rows per table is capped so
        that $n\,(p{+}1)\le 16{,}384$ cells; at $p{=}64$ this still allows the
        full $n{=}512$ rows. Gradients are accumulated over two half-batches to
        fit the widest tables in $40$\,GB of GPU memory. Training took $77.7$\,h
        on one A100 ($3.3\times$ the $p{\le}15$ model, whose cost scales roughly
        linearly in $p$).
  \item \textbf{FADO $p{\le}64$, fixed density.} Identical, except that the
        graph prior holds the expected number of parents per node fixed as $p$
        grows. The prior of \cref{tab:scm-priors} draws each earlier-to-later
        edge independently with probability $0.35$ ($0.40$ for the MLP and GP
        families), so the expected in-degree is about $2.5$ at $p{=}15$ but
        $5.5$ at $p{=}32$ and $11$ at $p{=}64$, and the fraction of features
        with a nonzero effect stays near $0.4$. For $p>15$ this variant scales
        every family's edge probability by $14/(p{-}1)$ (to $0.158$ at $p{=}32$
        and $0.078$ at $p{=}64$), which keeps the expected in-degree at its
        $p{=}15$ value, as TabPFN and TabICL~\citep{qu2025tabicl} do by holding
        graph density independent of width. For $p\le 15$ the prior is
        unchanged. Training took $36$\,h on one L40S.
\end{itemize}

\Cref{tab:wide-tables} evaluates all three models on $500$ held-out
mixture-prior SCMs at $p\in\{15,32,64\}$ with $n{=}512$ rows and the
heavy-tailed per-column noise of \cref{app:noise-families}, under both the
paper's graph density and the fixed density. The same $500$ SCM draws are used
for every model within a setting (fixed seed), so columns are paired. The
$p{\le}15$ model is applied unpadded at $p>15$; it has never seen a table wider
than $15$ columns.

Training on tables of up to $64$ features does not reduce accuracy at
$p{=}15$, and the resulting model beats the $p{\le}15$ model at every wider
setting (\cref{tab:wide-tables}). The $p{\le}15$ model still works beyond its
training width: at $p{=}64$ on fixed-density graphs it reaches Pearson $r$ of
$0.69$, against $0.45$ for the marginal baseline. Under the paper's density,
$p{=}64$ is hard for every model: each node has eleven parents in expectation,
and the best model barely beats predicting zero in MSE. On fixed-density
graphs of the same width, both models recover most of their $p{=}15$
accuracy. A padded $p\in[3,32]$ control, trained with the main text's recipe,
does about as well at $p{=}32$ under the paper's density ($r{=}0.64$ against
$0.62$ for the $p{\le}64$ model).

Holding the graph density fixed during training did not help. The
fixed-density model is worse than the $p{\le}64$ model in every row of
\cref{tab:wide-tables}, including at $p{=}15$, where the two training priors
are identical ($r$ of $0.73$ against $0.82$; seed-to-seed variation of the
$p{\le}15$ model is about $0.005$), and it is worse than the $p{\le}64$ model
even on the fixed-density evaluation graphs it was trained on ($0.68$ against
$0.76$ at $p{=}64$). One possible explanation is that a sparse wide table
has few nonzero effects, so most of its labels are zero and it carries less
training signal. The browser demo (\cref{app:code}) uses
the $p{\le}64$ model trained on the paper's density.

\begin{table}[h]
  \centering
  \small
  \caption{Wider tables: the paper's FADO model ($p\in[3,15]$) and two models
    trained on $p\in[3,64]$, evaluated on $500$ held-out mixture-prior SCMs per
    row block ($n{=}512$, heavy-tailed noise, seed-matched draws across
    columns). ``Density'' is the graph prior used to draw the evaluation SCMs:
    ``paper'' is the edge probability of \cref{tab:scm-priors}; ``fixed'' holds
    the expected in-degree at its $p{=}15$ value (identical to ``paper'' at
    $p{=}15$). ``Signal frac.'' is the fraction of features with nonzero
    $\tau$. The baseline column is the marginal-correlation reference of
    \cref{tab:all-baselines} for Pearson $r$ and AUROC, and the predict-zero
    estimator for MSE.}
  \label{tab:wide-tables}
  \input{tables/wide_tables}
\end{table}

\FloatBarrier
\section{Sachs Published DAG}
\label{app:sachs}

We take the DAG structure ($11$ nodes, $17$ edges) from
\citet{sachs2005causal}'s protein-signaling network and populate its edges
with $30$ random draws of linear non-Gaussian coefficients. Across all feature
draws, every model variant beats marginal correlation
(\cref{tab:sachs-ckpts}).

\begin{table}[h]
  \centering
  \caption{Sachs published DAG, $30$ random-coefficient draws.
    Pearson $(\tauhat,\tau)$ across all feature draws by FADO model variant;
    marginal and multivariate-linear baselines do not depend on the FADO
    variant (same data each row). Subscripts give paired-bootstrap $95\%$
    confidence intervals over the $30$ draws ($5{,}000$ resamples). The
    $\Delta$ column is the paired difference FADO~$-$~Marginal computed
    on each resample, so its CI accounts for the within-draw correlation
    between methods.}
  \label{tab:sachs-ckpts}
  \input{tables/sachs_ckpts}
\end{table}

\section{Architecture Ablations}
\label{app:ablations}

We retrained the foundation model from scratch three times, each time
removing one component and keeping every other setting ($150$k steps,
same prior, seed $0$), and evaluated each like the full model: $200$
held-out linear SCMs at $p{=}5$ padded to $15$ columns.

Without feature axis attention or the type embedding (the only signal
distinguishing the $Y$ column from feature columns), the model learns
nothing; without sample axis attention it still learns, but less
accurately (\cref{tab:ablations}).

\emph{Inference-time type-flip diagnostic}
        (\cref{tab:type-flip-ablation}): the retrain ablation could in
        principle reflect the model relying on the fixed position-$p$
        readout convention rather than on the type embedding signal.
        Holding the foundation model fixed, we re-evaluate
        on the same held-out set after substituting the type-id vector
        passed to the type-embedding lookup. With all-zero or mis-pointed type IDs, Pearson $r$ collapses
        (\cref{tab:type-flip-ablation}). The position-$p$ convention
        carries no signal at inference time, so we concluded that
	the trained model
        relies on the type embedding to identify the target column.

\begin{table}[h]
  \centering
  \caption{Architecture ablations. Each row removes one component
  from the FADO foundation model and retrains from scratch with the
  same recipe. All rows are evaluated on the same $200$ held-out linear
  SCMs ($p{=}5$ padded to $15$, heavy-tailed noise). Change in Pearson
  is relative to the full model. Pearson $r$ subscripts are Fisher-$z$
  $95\%$ CIs at $N{=}n_{\mathrm{scms}}$. Each ablation's CI is disjoint
  from the full model's.}
  \label{tab:ablations}
  \input{tables/ablations}
\end{table}

\begin{table}[h]
  \centering
  \caption{Inference-time type-id substitution on the trained foundation
    model (no retraining). $500$ held-out random SCMs,
    $n_{\mathrm{rows}}=512$. ``Mispointed'' = type-id vector with col $0$
    marked as target and col $p$ left as feature. ``All zero'' = no
    column marked. The position-$p$ readout convention is preserved in
    all rows; only the type-embedding input changes.}
  \label{tab:type-flip-ablation}
  \input{tables/type_flip}
\end{table}

\section{Motif Specifications and Results}
\label{app:motifs}

The five handcrafted DAG motifs used in \cref{sec:held-out-scms} stress
prototypical predictive-vs-causal failure modes: each motif has a
``key feature'' (coloured in \cref{fig:pred-vs-causal}) which is
non-causal but predictive in B/D/E (so $\tau=0$ despite high
marginal correlation with $Y$), and both causal and predictive in A/C.
We evaluate FADO and baselines on $50$ random-coefficient instances
of each motif, reporting the mean $\tauhat$ across instances with
paired-bootstrap $95\%$ confidence intervals (\cref{tab:motifs}).
FADO correctly assigns near-zero $\tauhat$ to the non-causal but
predictive features in B, D, and E, while marginal correlation
reports a strong effect on these features. Multivariate regression
and SHAP succeed on B, where conditioning on the observed confounder
$Z$ removes the spurious signal, but they fail on D and E, where the
spurious feature is downstream of $Y$ and not in the conditioning set.

\begin{table}[h]
  \centering
  \caption{Five motif data-generating processes.
    ``$B$ entries'' are structural-matrix coefficients; each motif samples
    edge weights from $\mathcal{U}(0.5,1.5)$ with random sign unless noted.}
  \label{tab:motif-specs}
  \begin{tabularx}{\linewidth}{l l X}
    \toprule
    Motif                     & DAG                           & Expected $\tau$       \\
    \midrule
    A: direct cause           & $X_1\!\to\!Y$                 & $\tau_1 > 0$          \\
    B: observed confounder    & $Z\!\to\!X_1,\;Z\!\to\!Y$     & $\tau_1=0$            \\
    C: mediator chain         & $X_1\!\to\!X_2\!\to\!Y$       & $\tau_1,\tau_2>0$     \\
    D: target descendant      & $Y\!\to\!X_1$                 & $\tau_1=0$            \\
    E: collider               & $X_1\!\to\!C\!\leftarrow\!Y$  & $\tau_C=0$            \\
    \bottomrule
  \end{tabularx}
\end{table}

\begin{table}[h]
  \centering
  \caption{Motif stress tests ($50$ random-coefficient instances per
    row, edge weights drawn from $\mathcal{U}(0.5, 1.5)$ with fixed
    positive sign; FADO foundation model). The SHAP column is
    path-dependent TreeSHAP via LightGBM (\cref{sec:baselines}),
    signed by marginal correlation and rescaled by $\mathrm{std}(Y)$.
    Bold cells: model correctly zeros a non-causal but predictive
    feature. Subscripts on Model $\tauhat$ are paired-bootstrap $95\%$
    CIs over the $50$ instances ($5{,}000$ resamples).}
  \label{tab:motifs}
  \resizebox{\linewidth}{!}{\input{tables/motifs}}
\end{table}

\section{Uncertainty: Coverage and Estimator Comparison}
\label{app:uncertainty}

\begin{table}[h]
  \centering
  \caption{Bootstrap-interval coverage on held-out SCMs: empirical coverage of
    nominal $50/80/90/95\%$ intervals (all features), coverage at $95\%$ split by
    true-cause vs.\ null features, and Spearman $\rho(\sigma,|\mathrm{err}|)$.
    The intervals are not calibrated, under-covering throughout. Coverage is
    lowest for true causes in the non-identifiable linear regime ($0.25$ at
    nominal $95\%$). The Spearman correlation between $\sigma$ and absolute
    error is always at least $0.66$.}
  \label{tab:uncertainty-coverage}
  \resizebox{\linewidth}{!}{\input{tables/uncertainty_coverage}}
\end{table}

\begin{table}[h]
  \centering
  \caption{Uncertainty quality by estimator: Spearman $\rho(\sigma,|\mathrm{err}|)$
    and selective-prediction gain for single-model bootstrap $\sigma$, a $K{=}5$
    deep ensemble, and CausalPFN's native posterior uncertainty (its per-column
    protocol; see \cref{app:offlabel-pfn}).}
  \label{tab:uncertainty-quality}
  \resizebox{\linewidth}{!}{\input{tables/uncertainty_quality}}
\end{table}

\section{Causal Chambers: Real Physical Interventions}
\label{app:chambers}

We evaluate on the light tunnel device of the Causal Chambers \citep{gamella2025chamber}. This physical system is accompanied by a causal graph, in which every actuator can be independently set, with ground truth effects measured directly from experiments. We use the uniform reference experiment ($n{=}10{,}000$); ground truth $\tau$ is computed at $\mu\pm\sigma$ from the intervention data. \Cref{tab:chambers-perfeature} reports per-feature results for the $\mathrm{vis}_1$ configuration, whose sensor decoys (e.g.\ $\mathrm{ir}_2$, correlation $0.90$ with the outcome, true $\tau=0$) are near-deterministic functions of several actuators.

Because the chamber randomizes every actuator, there is no confounding and regression recovers the effects accurately. Thus, this method serves as a sanity check on whether methods can avoid attributing effects to non-causal passive sensors. TreeSHAP ranks a non-causal sensor $\mathrm{ir}_2$ most highly, while a backdoor adjustment using the published graph correctly suppresses it. FADO also assigns these sensors nonzero mass because they are almost deterministic functions of the actuators, but FADO's uncertainty measure flags exactly these columns (\cref{sec:uncertainty}).

\begin{table}[h]
  \centering
  \caption{Causal Chambers (light tunnel, $\mathrm{vis}_1$ config): per-feature
    $\tauhat$ vs.\ measured ground-truth $\tau$. $\dagger$ marks DAG-null sensor
    decoys (true $\tau=0$).}
  \label{tab:chambers-perfeature}
  \resizebox{\linewidth}{!}{\input{tables/chambers_perfeature}}
\end{table}

\section{Semi-Synthetic Real-$X$ (UCI covariates)}
\label{app:semisynth-uci}

In this experiment, we test FADO on real covariates $X$ with synthetic outcomes $Y = f(X_C) + \varepsilon$ generated by a known SCM with a designated causal subset of features $C$. We inject the same proxy / leak / noise decoys as in \cref{sec:real-rcts} to isolate the real-covariate-distribution effect from the real-outcome effect. We achieve this property since $\tau$ is known by construction and the held-out Pearson $r$ and decoy false-positive rates remain well-defined. We test five settings: \texttt{diabetes} and \texttt{california\_housing} each crossed with \{linear, mixed nonlinear\} outcome, plus \texttt{abalone} with an interaction outcome (30 SCM instances per setting). We evaluate FADO on each setting.

The results are summarized in \cref{tab:semisynth-realx}. In the linear-outcome settings, the multivariate regression baseline wins the Pearson $r$ column as expected since it's the correct model, but it collapses on the mixed nonlinear settings. Moreover, FADO assigns smaller spurious effects to the decoys than marginal correlation does, except on one noise column where both are near zero.

\begin{table}[t]
  \centering
  \caption{Semi-synthetic real-X. Pearson $r$ between $\tauhat$ and $\tau$
  across $30$ SCMs per row at $n_{\mathrm{rows}}=512$; Proxy and noise columns are
  mean $|\tauhat|$ on proxy / pure-noise decoys (lower is better,
  $0$ is truth). Multivariate regression beats Pearson $r$ on the linear
  settings but collapses on the nonlinear ones; FADO assigns smaller spurious effects to the
  decoys than marginal correlation in all but one cell. Bold as in \cref{tab:all-baselines}.}
  \label{tab:semisynth-realx}
  \resizebox{0.80\linewidth}{!}{\input{tables/semisynth_realx}}
\end{table}

\section{Architecture and Training Details}
\label{app:architecture}

\begin{table}[h]
  \centering
  \caption{SCM families in the training prior of the FADO foundation model,
  with their relative weights. ``Exact $\tau$'' = closed-form label;
  ``MC $\tau$'' = Monte Carlo estimate of $\tau_i(M)$ averaged over
  $n_{\mathrm{do}}{=}4096$ samples from the post-intervention SCM.
  Random DAGs use edge probability $0.35$ ($0.40$ for the random MLP,
  MLP $+$ hidden and Gaussian process families). The linear family and
  those derived from it draw signed edge magnitudes
  $|B_{jk}|\sim\mathcal{U}(0.5,2.0)$. Noise $\varepsilon_j$ is drawn per
  column as in \cref{app:noise-families}.}
  \label{tab:scm-priors}
  \renewcommand{\arraystretch}{1.3}
  \begin{tabularx}{\linewidth}{>{\raggedright\arraybackslash}p{2.7cm} c >{\raggedright\arraybackslash}X >{\raggedright\arraybackslash}p{2.1cm}}
    \toprule
    Family & Weight & Generator & Label \\
    \midrule
    \arrayrulecolor{black!25}
    Linear & $0.30$ & $X_j=\sum_{k<j}B_{jk}X_k+\varepsilon_j$ & Exact $\tau$ \\\hline
    Mixed type & $0.16$ & linear; each feature independently median-split to
      $\pm1$ with probability $0.30$ & Exact $\tau$ \\\hline
    Polynomial nonlinear & $0.13$ & $X_j=\sum_{k<j}B_{jk}X_k+\sum_{k<j}C_{jk}(X_k^2-\mathbb{E} X_k^2)+\varepsilon_j$
      & MC $\tau$ \\\hline
    Random MLP & $0.13$ & $X_j=f_j(X_{\mathrm{pa}(j)})+\varepsilon_j$, $f_j$ a random
      one-hidden-layer $\tanh$ MLP & MC $\tau$ \\\hline
    Hidden confounder & $0.13$ & linear with one latent parent of three observed
      nodes, marginalized at sample time & Exact $\tau$ from full DAG \\\hline
    MLP $+$ hidden confounder & $0.05$ & random MLP with the same latent structure
      & MC $\tau$ from full DAG \\\hline
    Binary treatment & $0.05$ & linear DAG over $X$;
      $T=\mathbf{1}\{\gamma^\top X+\varepsilon_T+b>0\}$, $b\sim\mathcal{U}(-2,2)$;
      $Y=\beta_T T+\beta^\top X+\varepsilon_Y$ & MC $\tau$ \\\hline
    Univariate polynomial & $0.05$ & $X_j=\sum_{k<j}a_{jk}(X_k^{d_{jk}}-\mathbb{E} X_k^{d_{jk}})+\varepsilon_j$,
      $d_{jk}\in\{1,2,3\}$ & MC $\tau$ \\\hline
    Additive spline & $0.05$ & $X_j=\sum_{k<j}f_{jk}(X_k)+\varepsilon_j$, each $f_{jk}$
      a random combination of five Gaussian bumps & MC $\tau$ \\\hline
    Gaussian process & $0.05$ & $X_j=f_j(X_{\mathrm{pa}(j)})+\varepsilon_j$,
      $f_j\sim\mathrm{GP}$ with RBF kernel on at most three parents & MC $\tau$ \\\hline
    Multi-level mixed type & $0.05$ & linear; each feature independently
      quantile-binned into $K\in\{2,\dots,5\}$ levels with probability $0.40$
      & Exact $\tau$ \\
    \arrayrulecolor{black}
    \bottomrule
  \end{tabularx}
\end{table}

\begin{itemize}
  \item Layers: each layer applies feature axis attention (within each
        row, the $p{+}1$ cells attend to each other) and then sample axis
        attention (within each column, the $n$ cells attend to each
        other), using post-norm transformer blocks with feed-forward
        factor $2$. Width, heads and number of layers for each model are
        in \cref{tab:checkpoints}.
  \item Value projection: single linear layer $\R\!\to\!\R^{d}$ applied
        per cell of $D\in\R^{n\times(p+1)}$.
  \item Type embedding: learned $\R^{d}$ vector added to the value
        embedding; feature columns share one type, the $Y$ column has
        the other.
  \item Readout: the final cell embeddings are averaged over samples
        within each column, then a linear layer $\R^d\!\to\!\R$ is
        applied at feature positions.
  \item Training cost on one H100: $\approx 0.37$\,s/step, $15$\,h for
        the $150$k steps of the foundation model.
\end{itemize}

\begin{table}[h]
  \centering
  \caption{FADO model variants used for the experimental tables. All variants use the
  axial architecture of \cref{sec:arch} and share AdamW, a cosine schedule, $n=512$
  rows per SCM, and seed $0$. The single-family variants use
  $d_{\mathrm{model}}=96$, $4$ heads, $3$ layers, batch $32$,
  lr $3\!\times\!10^{-4}$ with a $200$-step linear warmup, and Laplace noise
  (Gaussian for the Gaussian controls); ``Large'' variants use
  $d_{\mathrm{model}}=192$, $8$ heads, $4$ layers. FADO (last row) uses
  $d_{\mathrm{model}}=384$, $8$ heads, $6$ layers ($14.2$M parameters),
  lr $10^{-4}$ with a $5{,}000$-step warmup, and the per-column noise of
  \cref{app:noise-families}.}
  \label{tab:checkpoints}
  \begin{tabular}{lllrr}
    \toprule
    Tag                          & Prior family                      & $p$    & Steps  & Batch \\
    \midrule
    Linear $p{=}5$               & linear non-Gaussian               & $5$    & $15{,}000$ & $32$ \\
    Linear $p{=}5$, 50k          & linear non-Gaussian               & $5$    & $50{,}000$ & $32$ \\
    Linear $p{=}8$               & linear non-Gaussian               & $8$    & $15{,}000$ & $32$ \\
    Linear $p{=}13$              & linear non-Gaussian               & $13$   & $20{,}000$ & $32$ \\
    Linear $p{=}20$              & linear non-Gaussian               & $20$   & $30{,}000$ & $32$ \\
    Large $p{=}5$                & linear non-Gaussian               & $5$    & $30{,}000$ & $32$ \\
    Large $p{=}13$               & linear non-Gaussian               & $13$   & $30{,}000$ & $32$ \\
    Hetero linear                & linear non-Gaussian               & $3$--$15$ & $15{,}000$ & $32$ \\
    Polynomial $p{=}5$           & polynomial nonlinear              & $5$    & $20{,}000$ & $32$ \\
    Random MLP $p{=}5$           & random MLP                        & $5$    & $20{,}000$ & $32$ \\
    Hidden $p{=}5$               & hidden confounder                 & $5$    & $20{,}000$ & $32$ \\
    Hidden $p{=}13$              & hidden confounder                 & $13$   & $25{,}000$ & $32$ \\
    MLP $+$ hidden $p{=}5$       & MLP $+$ hidden                    & $5$    & $25{,}000$ & $32$ \\
    Mixed $p{=}5$                & mixed type                        & $5$    & $20{,}000$ & $32$ \\
    Mixed $p{=}13$               & mixed type                        & $13$   & $25{,}000$ & $32$ \\
    Gaussian control $p{=}5$     & linear Gaussian (non-identifiable) & $5$    & $15{,}000$ & $32$ \\
    Gaussian control $p{=}8$     & linear Gaussian (non-identifiable) & $8$    & $15{,}000$ & $32$ \\
    \midrule
    FADO: $p\!\in\![3,15]$     & mixture (11 families) & $3$--$15$ & $150{,}000$ & $8$ \\
    FADO $p{\le}64$ (\cref{app:wide-tables}) & mixture (11 families) & $3$--$64$ & $150{,}000$ & $8$ \\
    FADO $p{\le}64$, fixed density & mixture, fixed in-degree & $3$--$64$ & $150{,}000$ & $8$ \\
    \bottomrule
  \end{tabular}
\end{table}

\FloatBarrier
\subsection{Per-column noise families}
\label{app:noise-families}

The FADO foundation model draws the noise family independently per column. For each feature column $j$, a family is sampled
uniformly from the seven listed in \cref{tab:noise-families}, the raw
values are drawn with the indicated parameters, and the column is then
centred to mean $0$ and rescaled to unit sample variance. The post-hoc
rescaling matters for the families whose parametric variance depends on a
random shape parameter (signed log-normal, bimodal Gaussian mixture).
Across a $p$-feature SCM, different columns therefore typically have
different marginal shapes.

\begin{table}[h]
  \centering
  \caption{Per-column noise families used by the FADO foundation model.
  For each feature column independently, a family is drawn uniformly,
  then the column's noise is sampled with the indicated parameters and
  rescaled to unit empirical variance. Random parameters are redrawn
  per column.}
  \label{tab:noise-families}
  \begin{tabular}{lll}
    \toprule
    Family & Raw draw & Random parameters \\
    \midrule
    \texttt{laplace}                & $\mathrm{Laplace}(0,\,1/\sqrt 2)$ & --- \\
    \texttt{gaussian}               & $\mathcal{N}(0,1)$ & --- \\
    \texttt{student\_t3}            & $t_3$ & --- \\
    \texttt{lognormal\_signed}      & $\mathrm{LogNormal}(0,\sigma)$, sign-flipped w.p.\ $0.5$ & $\sigma\sim\mathcal{U}(0.4,\,0.9)$ \\
    \texttt{exp\_signed}            & $S\cdot\mathrm{Exp}(1)$, $S\in\{\pm1\}$ Rademacher & --- \\
    \texttt{gauss\_mixture\_bimodal} & $\tfrac12\mathcal{N}(-s,\,0.5^2)+\tfrac12\mathcal{N}(s,\,0.5^2)$ & $s\sim\mathcal{U}(1.0,\,2.5)$ \\
    \texttt{skewnormal}             & $\alpha\,\lvert U\rvert + V$, $U,V\sim\mathcal{N}(0,1)$ & $\alpha\sim\mathcal{U}(-1.5,\,1.5)$ \\
    \bottomrule
  \end{tabular}
\end{table}

\newpage
\section{Using Interventional PFNs as Per-Feature Causal Estimators}
\label{app:offlabel-pfn}

We compared our method to CausalPFN \citep{balazadeh2025causalpfn}
and Do-PFN \citep{robertson2025dopfn}. However, neither of these
models were designed for the estimation of feature intervention effects.
To apply them, we designate each feature as the treatment variable, one
by one, and ask the interventional PFNs to estimate a causal effect.
More specifically, for each setting and each feature column $j \in \{1,\ldots,p\}$:
\begin{itemize}
	\item median-binarize $X_j$ to a $\{0,1\}$ pseudo-treatment so it
        matches the binary-$T$ prior baked into both released models;
\item treat the remaining $p-1$ columns of $X$ as the conditioning
        set, which includes downstream descendants of $X_j$, violating
        the pre-treatment-covariates assumption of both priors;
\item for CausalPFN, call \texttt{ATEEstimator} on
        $(X_{\setminus j}, \tilde X_j, Y)$; for Do-PFN, swap $X_j$ into
        column~$0$ of $D^{\mathrm{ob}}$ and integrate the predicted
        conditional interventional distribution to obtain
        $\widehat{\mathrm{ATE}}_j = \E[Y\mid\dop(\tilde X_j{=}1)]
          - \E[Y\mid\dop(\tilde X_j{=}0)]$ averaged over rows (the
        canonical Do-PFN-CATE reduction); 
\item rescale to per-$\sigma$-$X_j$
        units for comparability with FADO's standardized contrast.
\end{itemize}

CausalPFN and Do-PFN perform similarly in this setting and require
similar overhead.  FADO returns all $p$ effects in a single
	forward pass. For timings, see \cref{app:wallclock}. Both the
CausalPFN and Do-PFN runs were performed on a single RTX~5000~Ada,
with total compute under $10$~minutes across all $8$ settings.

\begin{table}[h]
  \centering
  \caption{Comparison of CausalPFN, Do-PFN, and FADO, computing
	Pearson correlation $(\tauhat,\tau)$ on the same
    $8$ regimes as \cref{tab:all-baselines}, $100$ held-out SCMs per
    row, $n_{\mathrm{rows}}=512$. The CausalPFN and Do-PFN columns are
    per-column applications of methods designed for a single
    user-designated binary treatment (\cref{sec:related}); see the four
    adaptations enumerated above. FADO is highest on
    every row.}
  \label{tab:offlabel-pfn}
  \resizebox{\linewidth}{!}{\input{tables/causal_offlabel_pfn}}
  \\[0.4em]
  \footnotesize{$^\ddagger$CausalPFN and Do-PFN are meant to estimate the effect
    of a single, user-designated binary treatment (\cref{sec:related}), so
    we are testing them in a setting for which they were not designed, while
    FADO simultaneously estimates the impact of every feature of a table.}
\end{table}

\newpage
\section{Paired-$\Delta$ Confidence Intervals for the Baseline Comparison}
\label{app:paired-delta}
This appendix accompanies \cref{tab:all-baselines}.

In the main text, we only had space to include point estimates. Confidence
intervals are shown below.

\begin{table}[h]
  \centering
  \caption{Paired-$\Delta$ Pearson $r$, FADO minus each baseline, on the same
    $8$ regimes and SCM stream as \cref{tab:all-baselines}.
    $\Delta = \mathrm{Pearson}(\mathrm{FADO})
    - \mathrm{Pearson}(\mathrm{baseline})$ with $95\%$ percentile-bootstrap
    CIs from $5{,}000$ resamples of the SCM index, paired so both methods
    see the same resampled SCMs.
    $n_{\mathrm{scms}}=500$, $n_{\mathrm{rows}}=512$.}
  \label{tab:all-baselines-paired-ci}
  \resizebox{\linewidth}{!}{\input{tables/all_baselines_paired_ci}}
\end{table}

\section{Per-Call Wallclock Comparison}
\label{app:wallclock}

FADO, as an amortized method, returns all $p$ per-feature effects in a single forward pass
  ($\sim$1\,ms on GPU regardless of $p\in\{5,13\}$), while per-feature
  CATE estimators (DoubleML, Causal Forest) scale linearly in $p$ because
  each feature triggers an independent fit/inference. 
  LightGBM-SHAP, DoubleML, Causal Forest pays the per-dataset training cost on every new dataset. 

The FADO foundation model (last row of \cref{tab:checkpoints},
        $14.2$M parameters) trained in $\sim 15$\,h $18$\,min on a single
        NVIDIA H100 GPU ($\approx 367$\,ms/step over $150{,}000$ steps;
        three seeds at $55{,}077$\,s, $55{,}136$\,s and $55{,}116$\,s). The
        single-family variants in \cref{tab:checkpoints} ($15$k--$50$k
        steps) are far cheaper: the largest, Linear $p{=}5$ at $50$k steps,
        trained and was evaluated in $59$\,min on one H100.
Per-call timing reported in \cref{tab:wallclock} is the
        median of $20$ repetitions on fresh linear non-Gaussian SCMs at
        $n_{\mathrm{rows}}{=}512$ (Causal Forest at $p{=}13$ CPU was
        capped at $5$ repetitions for budget; all other cells use $20$).

``inference'' rows pay only a forward pass per dataset.
        ``fit $+$ explain'' rows include training the per-dataset
        estimator (LightGBM regressor; LinearDML / CausalForestDML
        per feature) on the input table before producing attributions.

At $p{=}13$, FADO on GPU is roughly $1{,}100\times$ faster than
        Causal Forest on CPU at producing the length-$p$ contrast vector.

The FADO cells time the foundation model
        ($d_{\mathrm{model}}{=}384$,
        $6$ layers, $8$ heads) with the input padded to $15$ columns,
        as in the synthetic evaluations, so the $p{=}5$ and $p{=}13$ costs
        are the same. The FADO CPU cells were measured on a different
        single-thread laptop from the baseline CPU rows.

\begin{table}[h]
  \centering
  \caption{Per-call wallclock to produce the length-$p$ vector of
    per-feature causal contrasts for one fresh dataset. The FADO row is
    amortized: foundation pretraining is paid once and reused across all
    datasets. ``Fit $+$ explain'' rows train a fresh estimator on each
    input table. CPU column is a single-thread laptop; GPU column is
    $1\times$ L40s. Dashes denote cells we did not measure (predictive
    and fit-then-explain methods are CPU-native).}
  \label{tab:wallclock}
  \resizebox{\linewidth}{!}{\input{tables/wallclock}}
\end{table}

\section{Cross-Family Generalization}
\label{app:cross-family}
\begin{figure}[h]
  \centering
  \includegraphics[width=\linewidth]{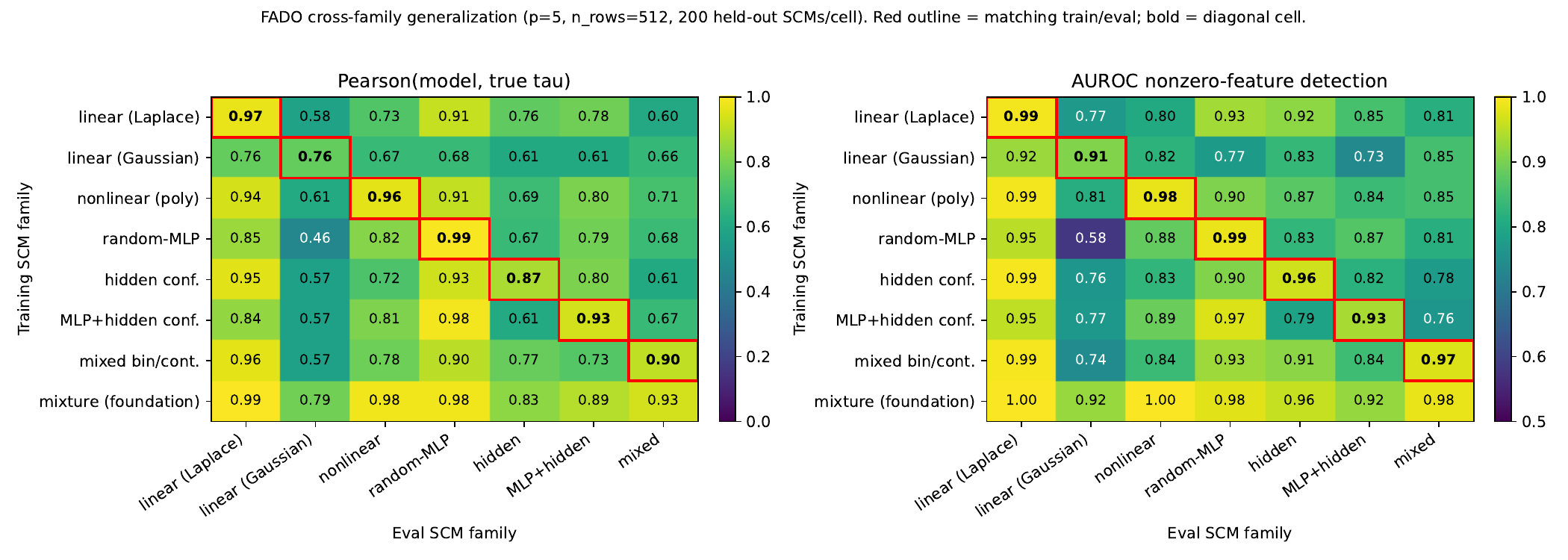}
  \caption{Cross-family generalization at $p{=}5$, $n_{\mathrm{rows}}{=}512$,
    $200$ held-out SCMs per cell. Rows are the training prior family
    (the $7$ single-family $p{=}5$ variants in \cref{tab:held-out-variants,tab:held-out-dags-extended}
    plus FADO); columns are the held-out eval
    family. Left: Pearson(model, true $\tau$). Right: AUROC for
    nonzero-feature detection. Red outlines and bold text mark cells
    where the training family matches the eval family; the foundation
    row has no diagonal cell because mixture is not an eval family.}
  \label{fig:cross-family}
\end{figure}

Our main FADO model is trained on a mixture prior. As a sensitivity analysis,
we investigated single-family ablations. In \cref{fig:cross-family} we
consider $8$ $p{=}5$ model variants (the $7$ single-family
        $p{=}5$ variants in \cref{tab:held-out-variants,tab:held-out-dags-extended} plus FADO).
	On the diagonal, where train and evaluation families match, performance
	is high, but off-diagonal performance does not drop too drastically.
	The mixture foundation FADO model outperforms single family models.
	The Gaussian evaluation column is the hardest, which makes sense
	from theory.

\section{SHAP Variants and the Predictor-to-SCM Gap}
\label{app:shap-variants}

\Cref{tab:target-vs-baselines} summarizes how the causal feature
profile $\tau$ relates to existing per-feature quantities used for
attribution and explanation: each estimator answers a different
question, and only $\tau$ is an interventional contrast.

The distinction between interventional and conditional SHAP is important
for explaining a fitted predictor, but it does not by itself identify
the SCM that generated the data. \citet{janzing2020feature}'s
``interventional SHAP is consistent with do-calculus'' claim is a claim
about the \emph{input distribution} of the predictor $f$: equivalently,
about a flat causal graph $\mathcal{G}_{\mathrm{flat}}$ in which every
feature is a root node and there are no edges between features. Under
$\mathcal{G}_{\mathrm{flat}}$, marginal sampling of $X_{-S}$ and
$\dop(X_S{=}x_S)$ coincide, so the marginal-baseline expectation
$\E[f(x_S, X_{-S})]$ is a well-defined do-quantity.

That flat graph is generally not the real SCM that produced the training
data. When the true mechanism has $X_2 = g(X_1)+\varepsilon$,
marginal-sampling $X_2$ while holding $X_1$ fixed can produce
$(X_1,X_2)$ pairs that occur in no observational or interventional
distribution under the SCM \citep{janzing2020feature, aas2021explaining}.
\citet{sundararajan2020manyshapley} compare the BShap, RBShap and CES
variants. Their analysis concerns the predictor's input model
and does not provide a way to recover SCM-level edges from
observations of $f$ alone. \citet{aas2021explaining} likewise frame the
interventional-vs-conditional choice as a question about feature
dependence in the observational distribution, not about identification
of the data-generating mechanism.

These variants all perform predictive attribution for the fitted model, rather than causally explaining the data-generating process. Shapley methods that do target the data-generating process, such as do-Shapley, require the causal graph (\cref{tab:shapley-comparison}). The choice among these variants also makes little difference in our experiments: path-dependent TreeSHAP, interventional TreeSHAP, and kernel SHAP agree closely on held-out SCMs (Pearson $r$ within $0.05$--$0.09$ of one another across five families), and all three assign the outcome leak their largest decoy score ($0.37$--$0.70$; true $\tau=0$). Counterfactual Shapley values \citep{lee2026improving} are defined with respect to a known SCM rather than a fitted predictor, so there is no data-only implementation to compare to. The
variants can be faithful to their respective predictive targets
($\mathcal{G}_{\mathrm{flat}}$ or the observational manifold) while
still answering a different question from the interventional feature
effect $\tau_i$.

\section{Hillstrom Hidden-Treatment Per-Feature Estimates}
\label{app:shap-contrast-table}

Full per-feature numbers from the Hillstrom hidden-treatment + decoys
experiment of \cref{sec:shap-contrast}, which uses the MineThatData e-mail
marketing dataset \citep{hillstrom2008}.

\begin{table}[h]
  \centering
  \caption{SHAP, DR-Learner, and FADO on Hillstrom hidden-treatment
  + decoys ($n{=}42{,}613$, visit outcome). DR-Learner
  \citep{kennedy2020drlearner} is run per feature with RandomForest
  first / second / final stages and 2-fold cross-fitting; binary
  features used directly as treatment indicators. Subscripts on the
  two $\tauhat$ columns are row-bootstrap $95\%$ CIs ($B{=}100$ for
  DR-Learner with the lighter RF config, $B{=}500$ for FADO).}
  \label{tab:shap-contrast}
  \resizebox{\linewidth}{!}{\input{tables/shap_contrast}}
\end{table}

\input{anchor-audit-appendix}

\input{proofs-appendix}

%% file: architecture-section.tex
\section{Architecture and Training}
\label{sec:architecture}

We train a transformer (\cref{sec:arch}) on a mixture of $11$ SCM prior families (\cref{sec:prior}) for $150{,}000$ steps with squared loss (\cref{sec:training}); the noise distribution of each column is drawn from seven families (\cref{app:noise-families}). The resulting foundation model, FADO (Feature-wise Amortized Do-Operators), takes datasets with between $3$ and $15$ features and extends to $64$ features in \cref{app:wide-tables}.

\subsection{Axial Transformer}
\label{sec:arch}

FADO treats observations as an exchangeable set, in the style of~\citet{pmlr-v97-lee19d}, and alternates attention along the two axes of the table, following~\citet{somepalli2022saint}. Given a dataset $D\in\R^{n\times(p+1)}$ where column $p+1$ is $Y$, FADO returns the full causal feature profile $\tauhat\in\R^p$ in a single forward pass. Each scalar cell is projected to a $d$-dimensional embedding, and a learned type embedding with two values (``feature'' and ``target'') is added to break the symmetry between features and target. The embeddings then pass through $6$ layers, each of which applies two post-norm multi-head attention blocks~\citep{vaswani2017attention}: feature axis attention, in which the $p+1$ cells of a row attend to each other, and sample axis attention, in which the $n$ cells of a column attend to each other. After the last layer, the cell embeddings are averaged over samples within each column, and a linear readout at the $p$ feature positions produces $\tauhat\in\R^p$. FADO uses $d{=}384$ and $8$ heads ($14.2$M parameters).

\begin{figure*}[t]
  \centering
  \includegraphics[width=0.75\linewidth]{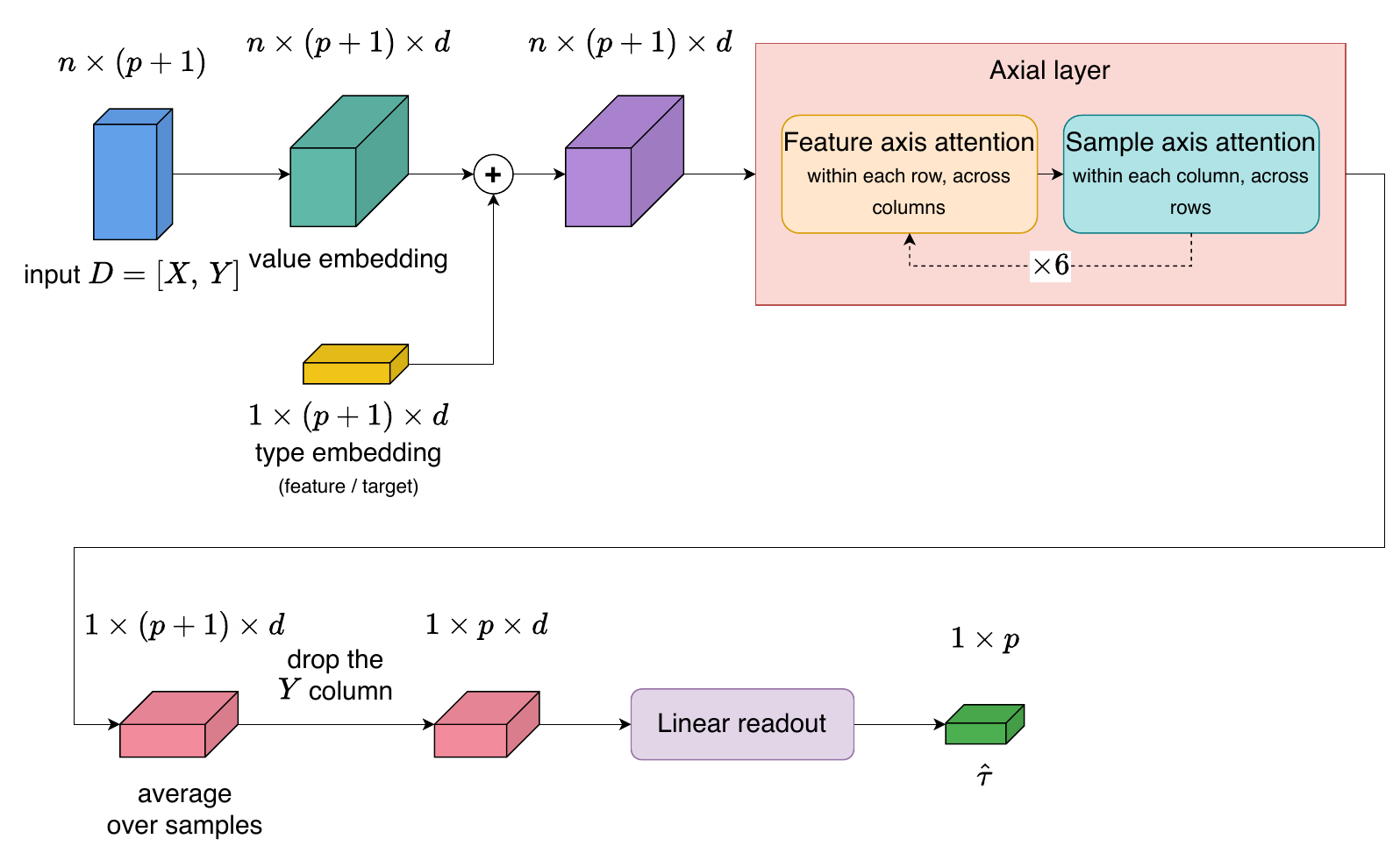}

  \caption{FADO architecture. Cell values are projected to $d$ dimensions and summed with a learned type embedding. Each of the $6$ layers applies feature axis attention within rows, then sample axis attention within columns. The cell embeddings are averaged over samples, and a linear head at the feature positions emits $\tauhat\in\R^p$.}
  \label{fig:arch}
\end{figure*}

\subsection{SCM Prior Families}
\label{sec:prior}

Each training SCM is drawn from one of $11$ prior families, summarized with their relative weights in \cref{tab:scm-priors} (\cref{app:architecture}). The families span linear, polynomial nonlinear, hidden confounder, and mixed-type settings. In the four families with linear mechanisms (linear, hidden confounder, and the two mixed-type families), the labels are computed in closed form (\cref{app:linear-tau}). As a special case, under no confounding (no open backdoor paths; ignorability), $\tau_i(M)=2\,\operatorname{Corr}(X_i,Y)$ (\cref{app:marginal-corr}). In the other seven families, $\tau_i(M)$ is estimated by Monte Carlo: $n_{\mathrm{do}}$ samples are drawn from the post-intervention SCM and averaged for each value of the intervention.

For each column $j$, the distribution of the noise $\varepsilon_j$ is drawn uniformly from seven families, independently across columns (\cref{app:noise-families}).

\subsection{Training Protocol}
\label{sec:training}

\paragraph{Data generation.} FADO is trained on synthetic datasets generated from the SCM priors described in \cref{sec:prior}. Each training step samples a batch of $B$ SCMs from the prior, simulates $n$ rows from each, and applies a random column permutation with the labels permuted accordingly. So that a single model can handle datasets with different numbers of features, datasets with fewer than $15$ features are padded to $15$ with columns of independent standard normal noise whose $\tau$ label is $0$, and the columns are permuted a second time so that the padding can appear at any position. At inference, datasets are padded the same way and the predictions at the padding columns are discarded.

\paragraph{Training objective.} The model parameters are optimized to minimize the squared loss on $\tauhat - \tau$. To improve training stability, we apply several guards: $\tau$ is clamped to $[-5,+5]$; batches with any non-finite label are skipped; and the gradient norm is clipped to $1.0$.

\paragraph{Optimizer and hyperparameters.} FADO is trained with the AdamW optimizer, a cosine schedule, learning rate $10^{-4}$ and a $5{,}000$-step linear warmup. For the single-family comparisons in \cref{sec:held-out-scms,app:cross-family} we also train smaller models, each on one prior family. These single-family variants use learning rate $3\!\times\!10^{-4}$ with a $200$-step warmup and Laplace noise in every column, except for two controls with Gaussian noise, where causal direction is not identifiable (\cref{sec:theory-identifiability}). Hyperparameters for all models are given in \cref{tab:checkpoints} (\cref{app:architecture}).

%% file: theory-section.tex
\section{Theory: What FADO Learns}
\label{sec:theory}

FADO is a prior-dependent amortized estimator of a causal functional in \cref{eq:tau-main}. The pretraining simulator defines a distribution over SCMs and provides intervention labels for each sampled SCM. At test time, the input is only an observational table, so the relevant population target is determined by the posterior over SCMs induced by the pretraining prior and that table. This section formalizes the target functional, shows that squared-loss pretraining identifies the posterior mean of the causal feature profile, and characterizes when that posterior mean converges to the true profile versus a prior-weighted average over observationally equivalent SCMs.

\subsection{Pretraining Prior and Target Functional}
\label{sec:theory-setup}

Let $\mathfrak{M}$ be a class of SCMs whose observed variables are
$(X_1,\ldots,X_p,Y)$, possibly together with additional latent variables. Let
$M\sim \Pi$ denote the pretraining prior over $\mathfrak{M}$.  The
observational law induced by $M$ on the observed variables is $P_M$, and an
observational table of $n$ rows is
\begin{equation}
  D_n=\{(x_r,y_r)\}_{r=1}^n \sim P_M^n .
\end{equation}

Assume $0<\Var_M(Y)<\infty$ throughout. The causal feature profile $\tau(M)=(\tau_1(M),\ldots,\tau_p(M))\in\mathbb{R}^p$ is defined component-wise by \cref{eq:tau-main}; the standardization puts all labels on the same outcome scale. We note that the causal feature profile is equivariant to relabeling of the feature columns (derivation in \cref{app:equivariance}), and the estimator of $\tau$ constructed in \Cref{sec:architecture} respects this equivariance property of the target by construction.

\subsection{Bayes-Optimal Amortized Predictor}
\label{sec:theory-bayes}

Consider the population squared-loss risk over measurable maps
$f:\mathcal{D}_n\to\mathbb{R}^p$,
\begin{equation}
  \mathcal{R}_2(f)
  =
  \E_{M\sim\Pi}
  \E_{D_n\sim P_M^n}
  \left[
    \|f(D_n)-\tau(M)\|_2^2
  \right].
  \label{eq:theory-risk}
\end{equation}

\begin{proposition}[Bayes action for the causal profile]
\label{prop:bayes-action}
Assume $\E_\Pi\|\tau(M)\|_2^2<\infty$.  Any population minimizer of
\cref{eq:theory-risk} satisfies
\begin{equation}
  f_\Pi^\star(D_n)
  =
  \E_{\Pi}\!\left[\tau(M)\mid D_n\right]
  \qquad \text{almost surely.}
  \label{eq:theory-posterior-mean}
\end{equation}
 Moreover, if the pretraining prior is invariant under relabeling of the feature columns, in the sense that $M\sim\Pi$ implies $M^\pi\sim\Pi$ for every permutation $\pi$, then
\begin{equation}
f_\Pi^\star(D_n^\pi)_j
=
f_\Pi^\star(D_n)_{\pi(j)},
\qquad
j=1,\ldots,p,\qquad \text{almost surely.}
\label{eq:equivariance-posterior-mean}
\end{equation}
\end{proposition}

\Cref{eq:theory-posterior-mean} shows that FADO is a PFN-style causal estimator
\citep{muller2022pfn,robertson2025dopfn,balazadeh2025causalpfn}: it is an amortized
estimator of the Bayesian posterior mean of the causal profile functional, where the posterior is induced by a pretraining prior.  It does not directly
recover ``the'' causal effect unless the posterior over SCMs assigns the same
causal feature profile to all observationally plausible SCMs. \Cref{eq:equivariance-posterior-mean} states that an invariant pretraining prior induces an equivariant posterior-mean target, which agrees with the equivariance of the true causal profile. Therefore, column order should not carry information during pretraining. We can enforce this by column-permuting the synthetic SCMs (and corresponding labels). This motivates the design choices in the transformer architecture (\Cref{sec:architecture}), in which computation is shared across feature columns and the model outputs one coordinate per feature, so that relabeling input features relabels output effects.

\subsection{Identifiability of the Feature Profile}
\label{sec:theory-identifiability}

Fix a data-generating SCM $M_0$ with observational law $P_0=P_{M_0}$ and
causal profile $\tau_0=\tau(M_0)$, and let $D_n\sim P_0^n$. Write
$\Pi_n(\cdot)=\Pi(\cdot\mid D_n)$. The feature profile is identifiable at $M_0$ under $\Pi$ if $\tau(M)=\tau_0$ for every observationally equivalent $M$ (i.e.\ every $M$ in the prior support with $P_M=P_0$). This is weaker than identifying the full graph or the full SCM: observationally equivalent SCMs may differ structurally while agreeing on the causal feature profile. Under three regularity conditions (formal versions in \cref{app:assumptions}) -- (A1) posterior consistency for the observational law, (A2) local continuity of $\tau(M)$ in $P_M$ at $P_0$, and (A3) a bounded-profile condition -- the posterior mean of $\tau(M)$ converges to $\tau_0$. (A2) is the key identifiability assumption.

\begin{proposition}[Consistency under local profile identifiability]
	\label{prop:profile-consistency}
	If \textup{(A1)--(A2)} hold, then, for every $\epsilon>0$,
		\begin{equation}
			\Pi_n\{M:\|\tau(M)-\tau_0\|_2>\epsilon\}
			\to 0
			\quad\text{in }P_0^n\text{-probability}.
			\label{eq:profile-posterior-concentration}
		\end{equation}

		If, in addition, \textup{(A3)} holds, then
		\begin{equation}
			f_{\Pi}^\star(D_n)
			=
			\E_\Pi[\tau(M)\mid D_n]
			\to
			\tau_0
			\quad\text{in }P_0^n\text{-probability}.
			\label{eq:profile-bayes-action-consistency}
		\end{equation}
\end{proposition}

\noindent
\Cref{eq:profile-posterior-concentration} states that learning the
observational law is sufficient to learn the causal feature profile when the
local continuity holds. \Cref{eq:profile-bayes-action-consistency}
then gives consistency of the posterior mean which in turn corresponds to the population optimizer of \cref{eq:theory-risk}. Note that the result concerns the population optimizer induced by the pretraining prior, it does not address the optimization error of a finite trained network. \Cref{app:theory-nscale} illustrates both regimes empirically on an identifiable prior (linear non-Gaussian) and a non-identifiable prior (linear Gaussian, where the direction of edges cannot be determined from observational data). \Cref{app:nonidentifiability} studies non-identifiable regimes showing that the posterior mean converges, under regularity, to the prior-weighted mean of the causal profiles compatible with $P_0$ and that the risk is lower bounded by the residual variation in $\tau(M)$ among observationally equivalent SCMs.

\paragraph{Where the prior helps.}
The identifiability status depends on the structural class.  In unrestricted
SCM classes, observational equivalence with different intervention effects is
common \citep{pearl2009causality,hernan2020causal}.  In purely Gaussian
linear SCMs, causal direction is generally not identified from observational
covariance alone.  In linear non-Gaussian acyclic models with independent
errors, the graph and total effects are identifiable under LiNGAM assumptions
\citep{shimizu2006lingam}.  In nonlinear additive-noise models, causal
direction is generically identifiable only under additional asymmetry
conditions \citep{hoyer2009anm}.  These results explain why a prior dominated
by non-Gaussian SCMs can learn a useful causal profile from observational
samples.  Gaussian and hidden-confounder SCMs fall outside these guarantees, so
we evaluate them separately.

\paragraph{Summary.}
FADO learns the Bayes action for $\tau(M)$ under the SCM pretraining prior,
which is the posterior mean $\E_\Pi[\tau(M)\mid D_n]$. With finite data, this
averages the causal profiles of SCMs weighted by their posterior probability
given the table. As $n\to\infty$, it converges to the true causal profile when
that profile is identifiable within the prior. Otherwise it converges to an
average over observationally equivalent SCMs weighted by the prior alone
(\cref{app:nonidentifiability}).

%% file: tables/foundation_per_family.tex
\begin{tabular}{lrrrr}
\toprule
 & \multicolumn{3}{c}{FADO $\tauhat$ vs $\tau$} & Marginal baseline \\
\cmidrule(lr){2-4} \cmidrule(lr){5-5}
Prior family & Pearson $r$ & AUROC (nonzero) & MAE (zero) & Pearson $r$ \\
\midrule
Linear & $0.95$ & $0.99$ & $0.02$ & $0.62$ \\
Mixed type & $0.89$ & $0.97$ & $0.07$ & $0.61$ \\
Polynomial nonlinear & $0.97$ & $0.99$ & $0.04$ & $0.60$ \\
Random MLP & $0.99$ & $0.97$ & $0.04$ & $0.68$ \\
Hidden confounder & $0.74$ & $0.88$ & $0.17$ & $0.60$ \\
MLP $+$ hidden confounder & $0.90$ & $0.90$ & $0.09$ & $0.62$ \\
Binary treatment & $0.95$ & $0.91$ & $0.17$ & $0.86$ \\
Univariate polynomial & $0.95$ & $0.94$ & $0.02$ & $0.60$ \\
Additive spline & $0.92$ & $0.83$ & $0.03$ & $0.67$ \\
Gaussian process & $0.85$ & $0.89$ & $0.03$ & $0.64$ \\
Multi-level mixed type & $0.90$ & $0.98$ & $0.05$ & $0.62$ \\
\bottomrule
\end{tabular}

%% file: tables/all_baselines.tex
\begin{tabular}{l c c c c c c c c}
\toprule
& \multicolumn{4}{c}{Linear} & \multicolumn{2}{c}{Nonlinear} & Hidden & Mixed \\
\cmidrule(lr){2-5} \cmidrule(lr){6-7}
& Gaussian & \multicolumn{3}{c}{Non-Gaussian} & Polynomial & MLP & confounder & type \\
& $p{=}5$ & $p{=}5$ & $p{=}8$ & $p{=}13$ & $p{=}5$ & $p{=}5$ & $p{=}5$ & $p{=}5$ \\
\midrule
\multicolumn{9}{l}{\emph{Causal discovery}} \\
\quad LiNGAM      & $0.39$ & $1.00$ & $\mathbf{0.99}$ & $\mathbf{0.97}$ & $0.57$ & $0.31$ & $0.76$ & $0.73$ \\
\quad NOTEARS     & $0.42$ & $0.40$ & $0.36$ & $0.31$ & $0.39$ & $0.51$ & $0.40$ & $0.39$ \\
\quad PC & $0.48$ & $0.47$ & $0.36$ & $0.16$ & $0.45$ & $0.44$ & $0.41$ & $0.44$ \\
\quad GES & $0.53$ & $0.51$ & $0.52$ & $0.30$ & $0.48$ & $0.47$ & $0.42$ & $0.47$ \\
\quad FCI & $0.25$ & $0.26$ & $0.23$ & $0.08$ & $0.28$ & $0.26$ & $0.24$ & $0.22$ \\
\addlinespace
\multicolumn{9}{l}{\emph{Treatment effect (per-column)$^\dagger$}} \\
\quad DML & $0.50$ & $0.52$ & $0.46$ & $0.38$ & $0.49$ & $0.57$ & $0.49$ & $0.52$ \\
\quad Causal Forest & $0.44$ & $0.44$ & $0.25$ & $0.13$ & $0.47$ & $0.58$ & $0.36$ & $0.48$ \\
\addlinespace
\multicolumn{9}{l}{\emph{Predictive attribution}} \\
\quad Marginal corr. & $0.62$ & $0.62$ & $0.59$ & $0.52$ & $0.58$ & $0.64$ & $0.55$ & $0.62$ \\
\quad Multivariate regression & $0.44$ & $0.44$ & $0.26$ & $0.13$ & $0.46$ & $0.59$ & $0.36$ & $0.47$ \\
\quad Permutation & $0.45$ & $0.43$ & $0.36$ & $0.25$ & $0.42$ & $0.62$ & $0.41$ & $0.44$ \\
\quad SHAP & $0.52$ & $0.51$ & $0.44$ & $0.34$ & $0.50$ & $0.64$ & $0.48$ & $0.53$ \\
\midrule
FADO & $\mathbf{0.79}$ & $0.98$ & $0.97$ & $0.89$ & $\mathbf{0.98}$ & $\mathbf{0.98}$ & $\mathbf{0.83}$ & $\mathbf{0.93}$ \\
\bottomrule
\end{tabular}

%% file: tables/acic_fp.tex
{\setlength{\tabcolsep}{4pt}
\begin{tabular}{lrrrrrrrrrrrr}
\toprule
& \multicolumn{3}{c}{Noise} & \multicolumn{3}{c}{Proxy} & \multicolumn{3}{c}{Leak} & \multicolumn{3}{c}{Conf-child} \\
\cmidrule(lr){2-4}\cmidrule(lr){5-7}\cmidrule(lr){8-10}\cmidrule(lr){11-13}
Family & F & Mg & Mv & F & Mg & Mv & F & Mg & Mv & F & Mg & Mv \\
\midrule
\multicolumn{13}{l}{\emph{No confounding}} \\
\quad Linear & $0.006$ & $0.025$ & $0.004$ & $0.533$ & $0.964$ & $0.013$ & $0.521$ & $1.934$ & $1.036$ & — & — & — \\
\quad Mixed nonlinear & $0.008$ & $0.022$ & $0.002$ & $0.230$ & $0.797$ & $0.008$ & $1.097$ & $1.986$ & $1.835$ & — & — & — \\
\quad Interaction & $0.004$ & $0.025$ & $0.005$ & $0.275$ & $0.711$ & $0.016$ & $0.464$ & $1.928$ & $1.610$ & — & — & — \\
\midrule
\multicolumn{13}{l}{\emph{Hidden confounder}} \\
\quad Linear & $0.004$ & $0.025$ & $0.003$ & $0.392$ & $0.795$ & $0.011$ & $0.350$ & $1.970$ & $1.331$ & $0.753$ & $1.041$ & $0.371$ \\
\quad Mixed nonlinear & $0.007$ & $0.021$ & $0.002$ & $0.182$ & $0.756$ & $0.008$ & $1.123$ & $1.990$ & $1.862$ & $0.344$ & $0.545$ & $0.059$ \\
\quad Interaction & $0.004$ & $0.026$ & $0.004$ & $0.197$ & $0.507$ & $0.014$ & $0.423$ & $1.965$ & $1.690$ & $0.813$ & $1.143$ & $0.225$ \\
\bottomrule
\end{tabular}}

%% file: tables/held_out_dags.tex
\begin{tabular}{lrrrr}
\toprule
 & \multicolumn{3}{c}{FADO $\tauhat$ vs $\tau$} & Marginal baseline \\
\cmidrule(lr){2-4} \cmidrule(lr){5-5}
Training setting & Pearson $r$ & AUROC (nonzero) & MAE (zero) & Pearson $r$ \\
\midrule
FADO & $0.99$ & $1.00$ & $0.01$ & $0.62$ \\
Linear $p{=}8$, 15k steps & $0.92$ & $0.98$ & $0.06$ & $0.59$ \\
Linear $p{=}13$, 20k steps & $0.82$ & $0.93$ & $0.12$ & $0.48$ \\
Linear $p{=}20$, 30k steps & $0.74$ & $0.89$ & $0.15$ & $0.43$ \\
Gaussian control $p{=}5$ & $0.77$ & $0.91$ & $0.16$ & $0.63$ \\
Polynomial nonlinear $p{=}5$ & $0.97$ & $0.99$ & $0.04$ & $0.62$ \\
Random MLP $p{=}5$ & $0.99$ & $0.98$ & $0.04$ & $0.64$ \\
Hidden confounder $p{=}5$ & $0.86$ & $0.95$ & $0.10$ & $0.57$ \\
Mixed type $p{=}5$ & $0.88$ & $0.97$ & $0.08$ & $0.62$ \\
\bottomrule
\end{tabular}

%% file: tables/held_out_dags_extended.tex
\begin{tabular}{lrrrr}
\toprule
 & \multicolumn{3}{c}{FADO $\tauhat$ vs $\tau$} & Marginal baseline \\
\cmidrule(lr){2-4} \cmidrule(lr){5-5}
Training setting & Pearson $r$ & AUROC (nonzero) & MAE (zero) & Pearson $r$ \\
\midrule
Linear $p{=}5$, 15k steps & $0.97$ & $0.99$ & $0.03$ & $0.64$ \\
Large $p{=}5$ ($d{=}192$, 4+4) & $0.99$ & $1.00$ & $0.01$ & $0.64$ \\
Large $p{=}13$ ($d{=}192$, 4+4) & $0.90$ & $0.97$ & $0.07$ & $0.48$ \\
Gaussian control $p{=}8$ & $0.75$ & $0.87$ & $0.19$ & $0.59$ \\
MLP + hidden confounder $p{=}5$ & $0.92$ & $0.93$ & $0.06$ & $0.62$ \\
Hidden confounder $p{=}13$ & $0.75$ & $0.91$ & $0.15$ & $0.47$ \\
Mixed type $p{=}13$ & $0.70$ & $0.87$ & $0.18$ & $0.48$ \\
\bottomrule
\end{tabular}

%% file: tables/seed_robust.tex
\begin{tabular}{lrrrr}
\toprule
Scale & Seed 0 & Seed 1 & Seed 2 & mean $\pm$ std \\
\midrule
$p{=}5$ & $0.951$ & $0.960$ & $0.945$ & $0.952 \pm 0.008$ \\
\bottomrule
\end{tabular}

%% file: tables/heterogeneous_p.tex
\begin{tabular}{rrrr}
\toprule
Eval $p$ & Heterogeneous & Dedicated & Gap \\
\midrule
$5$ & $0.954$ & $0.970$ & $-2$~pts \\
$8$ & $0.905$ & $0.924$ & $-2$~pts \\
$13$ & $0.788$ & $0.815$ & $-3$~pts \\
\bottomrule
\end{tabular}

%% file: tables/shapley_comparison.tex
\begin{tabular}{lrrrrrr}
\toprule
 & FADO & do-Shapley$^\dagger$ & Causal Shapley$^\dagger$ & DSCM$^\dagger$ & Marginal & Multivariate \\
\midrule
Linear & $\mathbf{0.99}$ & $0.95$ & $0.59$ & $0.97$ & $0.62$ & $0.41$ \\
Nonlinear & $\mathbf{0.98}$ & $0.97$ & $0.52$ & $0.95$ & $0.58$ & $0.37$ \\
MLP & $\mathbf{0.99}$ & $0.97$ & $0.77$ & $0.96$ & $0.67$ & $0.62$ \\
Hidden confounder & $0.80$ & $0.94$ & $0.55$ & $\mathbf{0.95}$ & $0.55$ & $0.34$ \\
MLP + hidden & $0.91$ & $\mathbf{0.98}$ & $0.68$ & $0.94$ & $0.57$ & $0.55$ \\
\bottomrule
\end{tabular}

%% file: tables/wide_tables.tex
\begin{tabular}{rllrrrrr}
\toprule
Eval $p$ & Density & Signal frac. & Metric & FADO & FADO $p{\le}64$ & FADO $p{\le}64$, fixed & Baseline \\
\midrule
\multirow{3}{*}{$15$} & \multirow{3}{*}{paper} & \multirow{3}{*}{$0.373$} & Pearson $r$ & $0.816$ & $0.819$ & $0.725$ & $0.542$ \\
 & &  & MSE & $0.151$ & $0.149$ & $0.215$ & $0.452$ \\
 & &  & AUROC (nonzero) & $0.885$ & $0.889$ & $0.836$ & $0.634$ \\
\midrule
\multirow{3}{*}{$32$} & \multirow{3}{*}{paper} & \multirow{3}{*}{$0.393$} & Pearson $r$ & $0.564$ & $0.623$ & $0.492$ & $0.340$ \\
 & &  & MSE & $0.474$ & $0.417$ & $0.517$ & $0.682$ \\
 & &  & AUROC (nonzero) & $0.796$ & $0.846$ & $0.764$ & $0.584$ \\
\midrule
\multirow{3}{*}{$32$} & \multirow{3}{*}{fixed} & \multirow{3}{*}{$0.263$} & Pearson $r$ & $0.776$ & $0.806$ & $0.702$ & $0.496$ \\
 & &  & MSE & $0.091$ & $0.076$ & $0.110$ & $0.217$ \\
 & &  & AUROC (nonzero) & $0.857$ & $0.886$ & $0.842$ & $0.676$ \\
\midrule
\multirow{3}{*}{$64$} & \multirow{3}{*}{paper} & \multirow{3}{*}{$0.391$} & Pearson $r$ & $0.274$ & $0.357$ & $0.214$ & $0.153$ \\
 & &  & MSE & $2.355$ & $2.210$ & $2.408$ & $2.524$ \\
 & &  & AUROC (nonzero) & $0.757$ & $0.838$ & $0.732$ & $0.590$ \\
\midrule
\multirow{3}{*}{$64$} & \multirow{3}{*}{fixed} & \multirow{3}{*}{$0.207$} & Pearson $r$ & $0.685$ & $0.759$ & $0.681$ & $0.454$ \\
 & &  & MSE & $0.078$ & $0.048$ & $0.059$ & $0.109$ \\
 & &  & AUROC (nonzero) & $0.804$ & $0.855$ & $0.815$ & $0.679$ \\
\bottomrule
\end{tabular}

%% file: tables/sachs_ckpts.tex
\begin{tabular}{l r r r r}
\toprule
Variant        & FADO & $\Delta$ vs Marginal corr. & Marginal corr. & Multivariate regression \\
\midrule
Linear $p{=}5$, 15k & $0.94\;[0.93,0.96]$ & $+0.06\;[+0.04,+0.08]$ & $0.88\;[0.86,0.90]$ & $0.67\;[0.65,0.70]$ \\
$p{=}8$, 15k & $0.94\;[0.92,0.95]$ & $+0.05\;[+0.03,+0.08]$ & $0.88\;[0.86,0.90]$ & $0.67\;[0.65,0.70]$ \\
$p{=}13$, 20k & $0.92\;[0.90,0.93]$ & $+0.04\;[+0.02,+0.06]$ & $0.88\;[0.86,0.90]$ & $0.67\;[0.65,0.70]$ \\
Foundation & $0.95\;[0.94,0.97]$ & $+0.07\;[+0.05,+0.09]$ & $0.88\;[0.86,0.90]$ & $0.67\;[0.65,0.70]$ \\
\bottomrule
\end{tabular}

%% file: tables/ablations.tex
\begin{tabular}{l r r r r}
\toprule
Configuration & Pearson & $R^2$ & AUROC (nonzero) & Change in Pearson \\
\midrule
FADO (all components) & $\mathbf{0.95\;[0.94,0.96]}$ & $0.904$ & $0.991$ & — \\
$-$ sample axis attention & $0.73\;[0.66,0.79]$ & $0.527$ & $0.887$ & $-23\%$ \\
$-$ feature axis attention & $0.00\;[-0.13,0.14]$ & $-0.000$ & $0.509$ & $-100\%$ \\
$-$ type embedding (feature/$Y$) & $-0.05\;[-0.18,0.09]$ & $-0.000$ & $0.482$ & $-105\%$ \\
\bottomrule
\end{tabular}

%% file: tables/type_flip.tex
\begin{tabular}{l r r r}
\toprule
Type-id config & Pearson($\hat\tau,\tau$) & MAE (zero) & Pearson($\hat\tau,\hat\tau_{\mathrm{normal}}$) \\
\midrule
Normal (col $p$ marked) & $0.991$ & $0.012$ & $1.000$ \\
All zero (no column marked) & $0.295$ & $0.105$ & $0.298$ \\
Mispointed (col $0$ marked) & $-0.000$ & $0.299$ & $-0.001$ \\
\bottomrule
\end{tabular}

%% file: tables/motifs.tex
\begin{tabular}{l l r r r r r}
\toprule
Motif (structure) & Key feature & Truth & Model $\tauhat$ & Marginal corr. & Multivariate regression & SHAP \\
\midrule
A: direct cause $X_1\!\to\!Y$ & $X_1$ & $1.33$ & $0.66\;[0.60,0.73]$ & $1.34$ & $1.34$ & $0.52$ \\
B: confounded $Z\!\to\!X_1,\,Z\!\to\!Y$ & $X_1$ & $0$ & $\mathbf{0.01\;[0.00,0.02]}$ & $1.15$ & $0.00$ & $0.07$ \\
C: mediator $X_1\!\to\!X_2\!\to\!Y$ & $X_1/X_2$ & $1.08/1.58$ & $0.56\;[0.50,0.62]$/$1.10\;[0.98,1.21]$ & $1.08/1.58$ & $-0.00/1.58$ & $0.08/0.60$ \\
D: target descendant $Y\!\to\!X_1$ & $X_1$ & $0$ & $\mathbf{-0.00\;[-0.00,-0.00]}$ & $1.36$ & $1.37$ & $0.52$ \\
E: collider $X_1\!\to\!C\!\leftarrow\!Y$ & $C$ & $0$ & $\mathbf{0.00\;[0.00,0.00]}$ & $1.11$ & $1.64$ & $0.53$ \\
\bottomrule
\end{tabular}

%% file: tables/uncertainty_coverage.tex
\begin{tabular}{lrrrrrrr}
\toprule
& \multicolumn{4}{c}{Empirical coverage (all features), nominal} & \multicolumn{2}{c}{Coverage@95} & \\
\cmidrule(lr){2-5}\cmidrule(lr){6-7}
Regime & $50$ & $80$ & $90$ & $95$ & Cause & Null & $\rho(\sigma,\mathrm{err})$ \\
\midrule
Linear (identifiable) & $0.403$ & $0.667$ & $0.758$ & $0.819$ & $0.833$ & $0.814$ & $0.884$ \\
Linear (non-ident.) & $0.165$ & $0.285$ & $0.337$ & $0.388$ & $0.246$ & $0.447$ & $0.889$ \\
Hidden conf. (non-ident.) & $0.350$ & $0.568$ & $0.669$ & $0.716$ & $0.726$ & $0.713$ & $0.890$ \\
MLP & $0.197$ & $0.377$ & $0.481$ & $0.560$ & $0.703$ & $0.674$ & $0.657$ \\
Mixture & $0.276$ & $0.472$ & $0.577$ & $0.651$ & $0.710$ & $0.728$ & $0.775$ \\
\bottomrule
\end{tabular}

%% file: tables/uncertainty_quality.tex
\begin{tabular}{lrrrrrr}
\toprule
& \multicolumn{2}{c}{FADO bootstrap} & \multicolumn{2}{c}{$K{=}5$ ensemble} & \multicolumn{2}{c}{CausalPFN native} \\
\cmidrule(lr){2-3}\cmidrule(lr){4-5}\cmidrule(lr){6-7}
Regime & $\rho$ & sel.\ gain & $\rho$ & sel.\ gain & $\rho$ & sel.\ gain \\
\midrule
Linear (ident.) & $0.896$ & $0.209$ & $0.864$ & $0.183$ & $-0.104$ & $-0.100$ \\
Linear (non-ident.) & $0.885$ & $0.416$ & $0.853$ & $0.398$ & — & — \\
Hidden conf. & $0.894$ & $0.338$ & $0.885$ & $0.322$ & $0.003$ & $-0.054$ \\
MLP & $0.648$ & $0.111$ & $0.516$ & $0.090$ & $-0.091$ & $-0.120$ \\
Mixture & — & — & — & — & — & — \\
\bottomrule
\end{tabular}

%% file: tables/chambers_perfeature.tex
\begin{tabular}{lrrrrr}
\toprule
Feature & $\tau_{\mathrm{true}}$ & FADO & SHAP & Marginal & Backdoor \\
\midrule
red & $0.515$ & $0.074$ & $0.127$ & $0.523$ & $0.523$ \\
green & $0.980$ & $0.463$ & $0.374$ & $0.929$ & $0.929$ \\
blue & $1.701$ & $0.658$ & $0.688$ & $1.584$ & $1.584$ \\
pol\_1 & $0.036$ & $0.012$ & $0.004$ & $0.015$ & $0.015$ \\
pol\_2 & $0.050$ & $0.001$ & $0.004$ & $0.017$ & $0.017$ \\
l\_11 & $-0.011$ & $0.003$ & $0.004$ & $0.009$ & $0.009$ \\
l\_12 & $-0.015$ & $0.008$ & $-0.007$ & $-0.014$ & $-0.014$ \\
current$^\dagger$ & $0.000$ & $0.422$ & $0.009$ & $1.345$ & $-0.006$ \\
angle\_1$^\dagger$ & $0.000$ & $0.012$ & $0.004$ & $0.014$ & $-0.298$ \\
ir\_2$^\dagger$ & $0.000$ & $0.922$ & $0.136$ & $1.375$ & $0.031$ \\
\bottomrule
\end{tabular}

%% file: tables/semisynth_realx.tex
\begin{tabular}{l r r r r r r r}
\toprule
 & \multicolumn{3}{c}{Pearson $(\tauhat,\tau)$} & \multicolumn{2}{c}{Proxy $|\tauhat|$} & \multicolumn{2}{c}{Noise $|\tauhat|$} \\
\cmidrule(lr){2-4}\cmidrule(lr){5-6}\cmidrule(lr){7-8}
Config & FADO & Marginal corr. & Multivariate regression & FADO & Marginal corr. & FADO & Marginal corr. \\
\midrule
diabetes / linear & $0.68$ & $0.50$ & $0.69$ & $\mathbf{0.54}$ & $0.96$ & $\mathbf{0.00}$ & $0.07$ \\
diabetes / mixed nonlin. & $\mathbf{0.70}$ & $0.45$ & $0.10$ & $\mathbf{0.33}$ & $0.64$ & $\mathbf{0.00}$ & $0.07$ \\
california / linear & $0.68$ & $0.52$ & $0.74$ & $\mathbf{0.35}$ & $0.97$ & $\mathbf{0.00}$ & $0.01$ \\
california / mixed nonlin. & $\mathbf{0.48}$ & $0.18$ & $0.03$ & $\mathbf{0.12}$ & $0.74$ & $0.02$ & $\mathbf{0.01}$ \\
abalone / interaction & $\mathbf{0.66}$ & $0.23$ & $0.43$ & $\mathbf{0.19}$ & $1.09$ & $\mathbf{0.01}$ & $0.03$ \\
\bottomrule
\end{tabular}

%% file: tables/causal_offlabel_pfn.tex
\begin{tabular}{l r r r}
\toprule
Setting & CausalPFN$^\ddagger$ & Do-PFN$^\ddagger$ & FADO \\
\midrule
Linear non-Gaussian $p{=}5$ & $0.476$ & $0.447$ & $0.99\;[0.99,1.00]$ \\
Linear non-Gaussian $p{=}8$ & $0.308$ & $0.355$ & $0.97\;[0.96,0.99]$ \\
Linear non-Gaussian $p{=}13$ & $0.153$ & $0.288$ & $0.86\;[0.81,0.92]$ \\
Linear Gaussian $p{=}5$ & $0.468$ & $0.478$ & $0.77\;[0.71,0.82]$ \\
Polynomial nonlinear $p{=}5$ & $0.628$ & $0.585$ & $0.98\;[0.97,0.99]$ \\
Random MLP $p{=}5$ & $0.678$ & $0.809$ & $0.98\;[0.96,0.98]$ \\
Hidden confounder $p{=}5$ & $0.463$ & $0.398$ & $0.85\;[0.80,0.89]$ \\
Mixed type $p{=}5$ & $0.498$ & $0.385$ & $0.93\;[0.88,0.96]$ \\
\bottomrule
\end{tabular}

%% file: tables/all_baselines_paired_ci.tex
\begin{tabular}{l c c c c c c c c}
\toprule
& \multicolumn{4}{c}{Linear} & \multicolumn{2}{c}{Nonlinear} & Hidden & Mixed \\
\cmidrule(lr){2-5} \cmidrule(lr){6-7}
& Gauss. & \multicolumn{3}{c}{Non-Gaussian} & Poly. & MLP & conf. & type \\
& $p{=}5$ & $p{=}5$ & $p{=}8$ & $p{=}13$ & $p{=}5$ & $p{=}5$ & $p{=}5$ & $p{=}5$ \\
\midrule
\multicolumn{9}{l}{\emph{Causal discovery}} \\
\quad LiNGAM & $0.40\,{\scriptstyle [0.36,0.44]}$ & $-0.01\,{\scriptstyle [-0.03,0.00]}$ & $-0.03\,{\scriptstyle [-0.04,-0.01]}$ & $-0.08\,{\scriptstyle [-0.10,-0.07]}$ & $0.41\,{\scriptstyle [0.36,0.46]}$ & $0.67\,{\scriptstyle [0.63,0.71]}$ & $0.07\,{\scriptstyle [0.04,0.11]}$ & $0.19\,{\scriptstyle [0.15,0.24]}$ \\
\quad NOTEARS & $0.37\,{\scriptstyle [0.33,0.42]}$ & $0.58\,{\scriptstyle [0.54,0.62]}$ & $0.60\,{\scriptstyle [0.57,0.64]}$ & $0.57\,{\scriptstyle [0.54,0.61]}$ & $0.59\,{\scriptstyle [0.54,0.63]}$ & $0.47\,{\scriptstyle [0.42,0.52]}$ & $0.43\,{\scriptstyle [0.39,0.48]}$ & $0.53\,{\scriptstyle [0.49,0.58]}$ \\
\quad PC & $0.31\,{\scriptstyle [0.26,0.36]}$ & $0.51\,{\scriptstyle [0.46,0.57]}$ & $0.61\,{\scriptstyle [0.57,0.65]}$ & $0.72\,{\scriptstyle [0.68,0.76]}$ & $0.53\,{\scriptstyle [0.48,0.59]}$ & $0.55\,{\scriptstyle [0.49,0.60]}$ & $0.42\,{\scriptstyle [0.36,0.48]}$ & $0.48\,{\scriptstyle [0.43,0.54]}$ \\
\quad GES & $0.26\,{\scriptstyle [0.19,0.33]}$ & $0.47\,{\scriptstyle [0.42,0.53]}$ & $0.45\,{\scriptstyle [0.40,0.50]}$ & $0.58\,{\scriptstyle [0.54,0.62]}$ & $0.50\,{\scriptstyle [0.45,0.55]}$ & $0.51\,{\scriptstyle [0.46,0.57]}$ & $0.42\,{\scriptstyle [0.36,0.47]}$ & $0.46\,{\scriptstyle [0.40,0.51]}$ \\
\quad FCI & $0.54\,{\scriptstyle [0.48,0.60]}$ & $0.73\,{\scriptstyle [0.67,0.79]}$ & $0.74\,{\scriptstyle [0.70,0.78]}$ & $0.80\,{\scriptstyle [0.71,0.82]}$ & $0.70\,{\scriptstyle [0.64,0.76]}$ & $0.73\,{\scriptstyle [0.67,0.78]}$ & $0.59\,{\scriptstyle [0.51,0.63]}$ & $0.70\,{\scriptstyle [0.65,0.76]}$ \\
\addlinespace
\multicolumn{9}{l}{\emph{Treatment effect (per-column)$^\dagger$}} \\
\quad DML & $0.29\,{\scriptstyle [0.25,0.32]}$ & $0.46\,{\scriptstyle [0.42,0.49]}$ & $0.50\,{\scriptstyle [0.47,0.53]}$ & $0.51\,{\scriptstyle [0.48,0.54]}$ & $0.49\,{\scriptstyle [0.45,0.52]}$ & $0.41\,{\scriptstyle [0.38,0.44]}$ & $0.34\,{\scriptstyle [0.30,0.37]}$ & $0.41\,{\scriptstyle [0.37,0.44]}$ \\
\quad Causal Forest & $0.35\,{\scriptstyle [0.30,0.39]}$ & $0.54\,{\scriptstyle [0.49,0.59]}$ & $0.71\,{\scriptstyle [0.67,0.76]}$ & $0.75\,{\scriptstyle [0.72,0.79]}$ & $0.50\,{\scriptstyle [0.46,0.54]}$ & $0.40\,{\scriptstyle [0.37,0.43]}$ & $0.47\,{\scriptstyle [0.42,0.51]}$ & $0.45\,{\scriptstyle [0.40,0.50]}$ \\
\addlinespace
\multicolumn{9}{l}{\emph{Predictive attribution}} \\
\quad Marginal corr. & $0.17\,{\scriptstyle [0.15,0.19]}$ & $0.36\,{\scriptstyle [0.34,0.39]}$ & $0.37\,{\scriptstyle [0.35,0.40]}$ & $0.37\,{\scriptstyle [0.34,0.39]}$ & $0.40\,{\scriptstyle [0.37,0.43]}$ & $0.34\,{\scriptstyle [0.31,0.37]}$ & $0.28\,{\scriptstyle [0.25,0.31]}$ & $0.31\,{\scriptstyle [0.28,0.34]}$ \\
\quad Multivariate regression & $0.35\,{\scriptstyle [0.30,0.39]}$ & $0.54\,{\scriptstyle [0.49,0.59]}$ & $0.71\,{\scriptstyle [0.66,0.75]}$ & $0.76\,{\scriptstyle [0.72,0.79]}$ & $0.52\,{\scriptstyle [0.48,0.55]}$ & $0.39\,{\scriptstyle [0.36,0.42]}$ & $0.47\,{\scriptstyle [0.42,0.51]}$ & $0.45\,{\scriptstyle [0.40,0.50]}$ \\
\quad Permutation & $0.34\,{\scriptstyle [0.30,0.37]}$ & $0.55\,{\scriptstyle [0.51,0.59]}$ & $0.61\,{\scriptstyle [0.56,0.65]}$ & $0.63\,{\scriptstyle [0.60,0.67]}$ & $0.55\,{\scriptstyle [0.51,0.59]}$ & $0.36\,{\scriptstyle [0.33,0.40]}$ & $0.42\,{\scriptstyle [0.37,0.48]}$ & $0.49\,{\scriptstyle [0.44,0.53]}$ \\
\quad SHAP & $0.27\,{\scriptstyle [0.24,0.30]}$ & $0.47\,{\scriptstyle [0.44,0.50]}$ & $0.53\,{\scriptstyle [0.50,0.56]}$ & $0.55\,{\scriptstyle [0.52,0.58]}$ & $0.48\,{\scriptstyle [0.44,0.51]}$ & $0.35\,{\scriptstyle [0.32,0.37]}$ & $0.35\,{\scriptstyle [0.32,0.40]}$ & $0.39\,{\scriptstyle [0.36,0.43]}$ \\
\bottomrule
\end{tabular}

%% file: tables/wallclock.tex
\begin{tabular}{l l r r r r}
  \toprule
  & & \multicolumn{2}{c}{$p{=}5$, $n{=}512$} & \multicolumn{2}{c}{$p{=}13$, $n{=}512$} \\
  \cmidrule(lr){3-4} \cmidrule(lr){5-6}
  Method & Cost type & CPU (s) & GPU (s) & CPU (s) & GPU (s) \\
  \midrule
  Marginal corr.           & inference        & $0.0004$ & ---       & $0.0004$ & ---       \\
  Multivariate regression  & inference        & $0.0001$ & ---       & $<\!0.0001$ & ---     \\
  LightGBM-SHAP            & fit $+$ explain  & $0.041$  & ---       & $0.080$   & ---       \\
  DoubleML (LinearDML)     & fit $+$ explain  & $0.99$   & ---       & $3.34$    & ---       \\
  Causal Forest            & fit $+$ explain  & $5.27$   & ---       & $22.95$   & ---       \\
  \textbf{FADO (ours)}     & inference (amortized) & $\mathbf{0.98}$  & $\mathbf{0.020}$ & $\mathbf{1.01}$ & $\mathbf{0.020}$ \\
  \bottomrule
\end{tabular}

%% file: tables/shap_contrast.tex
\begin{tabular}{l r r r r r}
\toprule
Feature & SHAP rank & $|\mathrm{SHAP}|$ & DR-Learner $\tauhat$ & FADO $\tauhat$ & Truth \\
\midrule
Leak $L_Y$ & \textbf{\#1} & $3.67$ & $\mathbf{+1.08\;[+1.06,+1.09]}$ & $\mathbf{+1.03\;[+0.67,+1.32]}$ & $0$ \\
Proxy $P_T$ & \#7 & $0.07$ & $+0.02\;[+0.01,+0.03]$ & $+0.16\;[+0.02,+0.36]$ & $0$ \\
noise$_1$ & --- & $0.07$ & $-0.00\;[-0.01,+0.01]$ & $-0.02\;[-0.08,+0.06]$ & $0$ \\
noise$_2$ & --- & $0.09$ & $-0.00\;[-0.01,+0.01]$ & $-0.02\;[-0.09,+0.05]$ & $0$ \\
noise$_3$ & --- & $0.07$ & $-0.00\;[-0.01,+0.01]$ & $-0.02\;[-0.09,+0.05]$ & $0$ \\
$T_{\mathrm{true}}$ & \#3 & $0.09$ & $+0.03\;[+0.02,+0.03]$ & $-0.01\;[-0.06,+0.09]$ & $+0.22$ RCT \\
newbie & \#2 & $0.13$ & $-0.03\;[-0.04,-0.03]$ & $-0.16\;[-0.34,-0.06]$ & real \\
history & \#5 & $0.08$ & $+0.03\;[+0.02,+0.04]$ & $+0.22\;[+0.05,+0.46]$ & real \\
recency & \#9 & $0.05$ & $-0.03\;[-0.04,-0.02]$ & $-0.05\;[-0.13,-0.00]$ & real \\
mens & --- & --- & $+0.01\;[+0.01,+0.02]$ & $-0.04\;[-0.11,+0.02]$ & real \\
womens & --- & --- & $+0.01\;[+0.01,+0.02]$ & $-0.05\;[-0.14,+0.02]$ & real \\
\bottomrule
\end{tabular}

%% file: anchor-audit-appendix.tex
\section{Anchor Audit: Is the $(-1, +1)$ Contrast a Faithful Summary?}
\label{app:anchor-audit}

FADO returns a single per-feature scalar $\tauhat_i$ trained to
match the symmetric population contrast
$\Delta_i(-1, +1) = \E[Y \mid \dop(X_i = +1)] - \E[Y \mid \dop(X_i = -1)]$.
This anchor pair has a known degeneracy: for any mechanism that is
even-symmetric in $X_i$ around the origin (e.g.\ $Y = X_i^2 + \varepsilon$),
$\Delta_i(-1, +1) \equiv 0$ even though intervening on $X_i$ does move
$Y$.  We check whether this affects five real datasets: Hillstrom, LaLonde,
Criteo, Sachs and Tübingen.

\paragraph{Method.}
Per feature $X_i$ in each dataset, no model retraining:

\begin{itemize}
  \item Standardize $X_i$ to zero mean / unit variance and $Y$ likewise.
  \item Estimate $\hat r_i(t) = \widehat{\E}[Y_{\rm std} \mid X_{i,\rm std} = t]$
        at the fixed grid $t \in \{-2, -1, 0, +1, +2\}$ via Nadaraya-Watson
        with a Gaussian kernel (bandwidth $0.4$ in standardized $X$ units),
        and an effective-sample-size guard $n_{\mathrm{eff}}(t) \geq 25$
        to keep heavy-tail outliers from dominating any single anchor.
  \item In parallel, compute a rank-based crosscheck
        $\hat r_i(F^{-1}(q))$ at $q \in \{0.1, 0.3, 0.5, 0.7, 0.9\}$.
        Each rank bin has equal sample count by construction, so the
        rank-based summary is robust to heavy tails.
  \item \textbf{Flag} feature $X_i$ when both
        $|\hat\Delta_i(-1, +1)| \leq 0.05$ \emph{and}
        $\max_q \hat r_i(F^{-1}(q)) - \min_q \hat r_i(F^{-1}(q)) \geq 0.20$.
        Two-sided in standardized $Y$ units.
\end{itemize}

The audit is observational.  $\hat r_i$ estimates
$\E[Y \mid X_i = t]$, not $\E[Y \mid \dop(X_i = t)]$.  Confounding
biases the symmetric contrast and the curve range identically, so the
\emph{comparison} is unaffected; a flagged feature is not necessarily a
flagged \emph{causal} effect.

\paragraph{Result.}
Across the five real datasets, $3$ of $149$
feature-level audits flag the symmetric-contrast blind spot
(\cref{tab:anchor-audit-summary}).

\begin{table}[h]
  \centering
  \caption{Anchor audit across five real datasets. Flagged
  features have $|\hat\Delta_i(-1, +1)| \leq 0.05$ and rank-based curve
  range $\geq 0.20$. Hillstrom and Criteo column counts include the
  signed treatment plus one-hot encodings of categorical covariates;
  Tübingen counts only the $99$ two-variable pairs.}
  \label{tab:anchor-audit-summary}
  \begin{tabular}{l r r r}
    \toprule
    Dataset    & $n$       & Audits & Flagged \\
    \midrule
    Hillstrom  & $64{,}000$ & $19$  & $0$ \\
    LaLonde    & $614$      & $9$   & $0$ \\
    Criteo     & $20{,}000$ & $13$  & $1$ (\texttt{f2}) \\
    Sachs      & $8{,}000$  & $10$  & $0$ \\
    Tübingen   & varies     & $99$  & $2$ (\texttt{pair0024}, \texttt{pair0094}) \\
    \midrule
    \textbf{Total} & — & $\mathbf{149}$ & $\mathbf{3}\;(\approx 2\%)$ \\
    \bottomrule
  \end{tabular}
\end{table}

The three flagged cases have non-monotone observational shapes.  Two are
inverted-U and one is a threshold at the negative tail:

\begin{itemize}
  \item \textbf{Criteo \texttt{f2}.} Rank-based curve
        $(-0.21, -0.21, +0.46, +0.13, -0.20)$.
        $|\hat\Delta_{f2}(-1, +1)| = 0.031$, rank range $0.68$.
        An asymmetric anchor pair $(+0.5, +1.5)$ already shows
        $-0.32$.  The curve drops above the median.  The covariate is anonymised, so we make no causal
        claim, only that its observational shape is not symmetric.
  \item \textbf{Tübingen \texttt{pair0024}.}
        Gaussian-anchor curve $(+0.59, -0.08, -0.12, -0.06, -0.05)$.
        Threshold-like at the negative tail: only the lowest-$X$
        bin produces a non-trivial response.
        $|\hat\Delta(-1, +1)| = 0.021$, rank range $0.27$.
  \item \textbf{Tübingen \texttt{pair0094}.}
        Gaussian-anchor curve $(-0.37, -0.09, +0.33, -0.07, -0.27)$.
        Inverted-U.  $|\hat\Delta(-1, +1)| = 0.015$, rank range $0.81$.
\end{itemize}

\begin{figure}[h]
  \centering
  \includegraphics[width=0.78\linewidth]{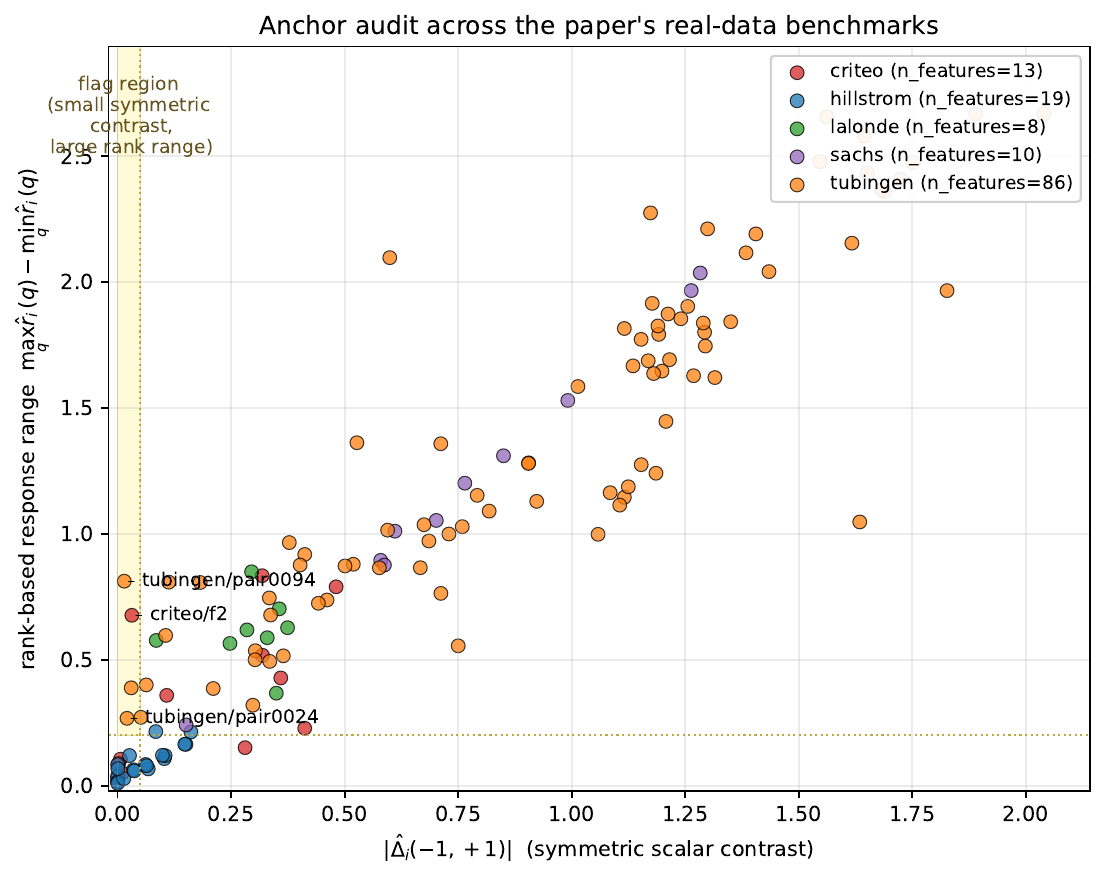}
  \caption{Per-feature $|\hat\Delta_i(-1,+1)|$ against the rank-based
  curve range across the five real datasets ($149$ feature
  audits).  Shaded strip on the left is the flag region (small
  symmetric contrast and large rank range).  Three features are flagged
  out of $149$.}
  \label{fig:anchor-audit-scatter}
\end{figure}

\Cref{fig:anchor-audit-scatter} plots
$|\hat\Delta_i(-1, +1)|$ against the rank-based curve range over all
$149$ feature audits, colour-coded by dataset; the flag region is the
shaded strip on the left.  Most features sit on or above the
$y = x$ line: symmetric contrast and rank-based range agree, i.e.\ the
effect is monotone and the symmetric pair captures it.  Sachs is the
clearest case.  Every protein lies far from the flag region because the
benchmark SCM is linear by construction.  Tübingen, despite being
\emph{curated} for nonlinear cause-effect relationships, mostly contains
\emph{monotone-nonlinear} pairs (saturating, log-link), which the
symmetric pair handles fine; only the non-monotone pairs are
flagged.

\begin{figure}[h]
  \centering
  \includegraphics[width=\linewidth]{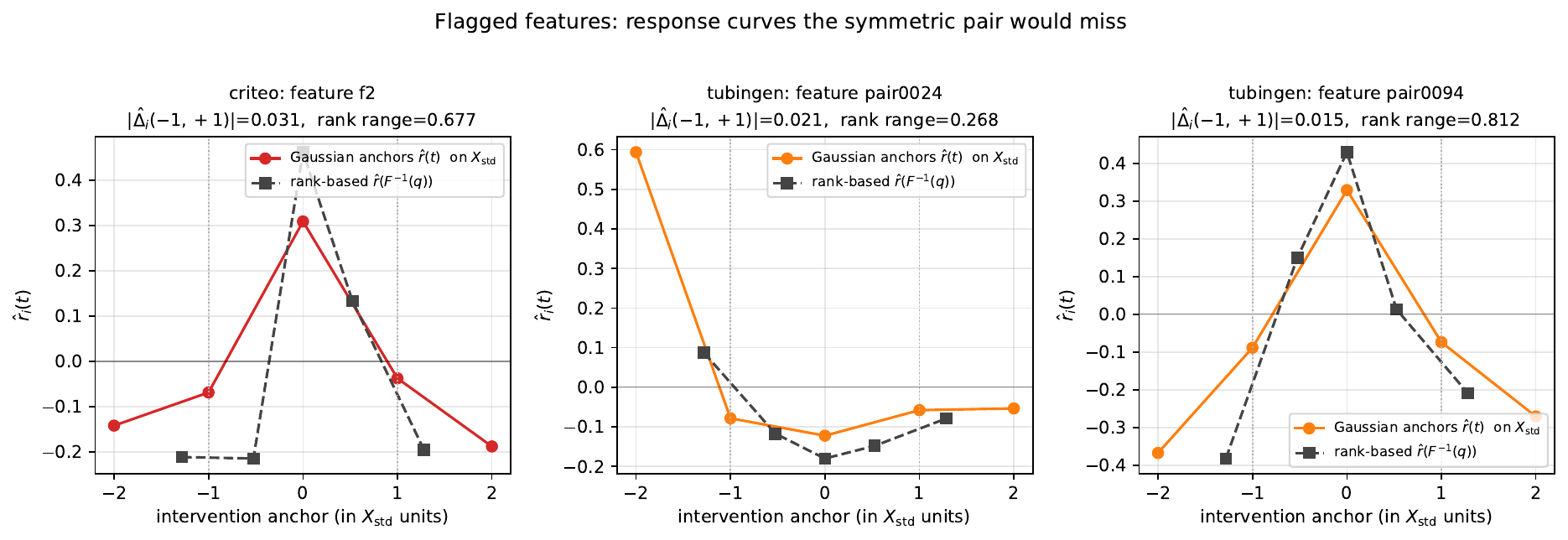}
  \caption{Response curves for the three flagged features.  Solid lines
  are Nadaraya-Watson estimates at the Gaussian-anchor grid $t \in
  \{-2,-1,0,+1,+2\}$ (in standardized $X$ units); dashed lines are the
  rank-based crosscheck at $q \in \{0.1,0.3,0.5,0.7,0.9\}$ plotted at
  the corresponding standard-normal quantiles.  The thin vertical
  dotted lines mark $\pm 1$, the only two anchors the symmetric pair
  reads.}
  \label{fig:anchor-audit-curves}
\end{figure}

\paragraph{Reading the result.}
This is mostly a negative result and we frame it as such.

\begin{itemize}
  \item The symmetric-contrast head $\Delta_i(-1, +1)$ is a defensible
        default on the datasets in this paper: $\approx 97\%$ of feature
        audits across $5$ datasets show no blind spot.
  \item The known degeneracy on symmetric nonlinear mechanisms does
        occur in practice, but rarely.  We found three instances in
        $149$ audits.  When it does occur, the pattern is a peak-and-decay or
        threshold shape.
  \item A natural extension is to widen the head from a single
        anchor pair to a $K$-anchor curve readout $\hat r_i(t_k)$ at
        $t_k \in \{-2, -1, 0, +1, +2\}$, of which the existing
        $\Delta_i(-1, +1)$ is one entry.  We do not pursue this in
        the present paper because the empirical incidence here is too
        low to motivate the extra parameters; we flag it as the right
        next step \emph{when} a target dataset shows
        Criteo-\texttt{f2}- or \texttt{pair0024}-style structure.
\end{itemize}

%% file: proofs-appendix.tex
\section{Further Details on Theoretical Results}
\label{app:proofs}

\subsection{Equivariance of the causal feature profile}
\label{app:equivariance}

The causal feature profile is equivariant under relabeling of feature columns. Let $\pi$ be a permutation of $\{1,\ldots,p\}$, and write $M^\pi$ and $D_n^\pi$ for the SCM and dataset with features relabeled by $X'_j=X_{\pi(j)}$, leaving the outcome $Y$ unchanged. Then
\[
\tau_j(M^\pi)
=
\frac{
	\mathbb{E}_{M}[Y\mid do(X_{\pi(j)}=b_{\pi(j)})]
	-
	\mathbb{E}_{M}[Y\mid do(X_{\pi(j)}=a_{\pi(j)})]
}{
	\sigma_Y(M)
}
=
\tau_{\pi(j)}(M).
\]
The estimator constructed in \cref{sec:architecture} respects this equivariance by construction.

\subsection{Linear continuous SCM closed-form label}
\label{app:linear-tau}

For the linear continuous SCM family, the causal feature profile admits a closed-form label. Let $Z=(Z_1,\ldots,Z_{p+1})$ collect the observed variables, with feature $X_i$ corresponding to node $z_i$ and outcome $Y$ corresponding to node $y$. Suppose
\begin{equation}
  Z = BZ + E,
\end{equation}
where $B_{jk}$ is the coefficient from parent $k$ to child $j$, and let $A=(I-B)^{-1}$. If the intervention levels $(a_i,b_i)$ are expressed in standardized units of $X_i$, then the standardized outcome contrast is
\begin{equation}
  \tau_i(M)
  =
  (b_i-a_i)\, A_{y,z_i}\,\frac{\sigma_{X_i}}{\sigma_Y} .
  \label{eq:linear-tau}
\end{equation}
For the families with nonlinear mechanisms, $\tau_i(M)$ is estimated by Monte Carlo draws from the post-intervention SCM (\cref{sec:prior}).

\subsection{Reduction to marginal correlation}
\label{app:marginal-corr}

We note that \cref{eq:linear-tau} reduces to marginal
correlation in the exogenous one-feature linear case.  Let $X_i$ and $Y$ be population-standardized to unit variance, with the contrast
$(a_i,b_i)=(-1,+1)$, and
\[
Y=\beta_i X_i+\eta_i,
\qquad
\mathbb{E}[X_i\eta_i]=0 .
\]
Equivalently, the only contribution to the covariance between $X_i$ and $Y$ is the
total linear effect of $X_i$ on $Y$; in the single-edge case
$X_i\to Y$, this total effect is the direct coefficient $\beta_i$.
Then $A_{y,z_i}=\beta_i$ and
$\operatorname{Corr}(X_i,Y)=\operatorname{Cov}(X_i,Y)=\beta_i$, so
\[
\tau_i(M)
=
2\,\operatorname{Corr}(X_i,Y).
\label{eq:theory-corr-reduction}
\]
We use marginal correlation as a baseline in \cref{sec:baselines}. It will
only recover the correct causal effect under the assumptions of exogeneity and
linearity, while it will fail under confounding, reversed direction,
target-descendant leakage, collider or selection effects, etc.

\subsection{Formal assumptions and technical preliminaries}
\label{app:assumptions}

This subsection collects formal definitions and assumptions referenced informally in \cref{sec:theory}. Notation matches that section: $M_0$ is a data-generating SCM with observational law $P_0=P_{M_0}$ and causal profile $\tau_0=\tau(M_0)$; $D_n\sim P_0^n$; $\Pi_n(\cdot)=\Pi(\cdot\mid D_n)$.

\paragraph{Bounded Lipschitz metric.}
We metrize weak convergence of observational laws on $\mathcal{Z}=\mathbb{R}^{p+1}$ by the bounded Lipschitz metric
\begin{equation}
	d_{\mathrm{BL}}(P,Q)
	=
	\sup_{\|h\|_\infty\le 1,\ \operatorname{Lip}(h)\le 1}
	\left|
	\int h\,dP-\int h\,dQ
	\right|.
	\label{eq:bl-metric}
\end{equation}

\paragraph{Effective prior support and equivalence classes.}
Let $\mathfrak{M}_\Pi$ denote an effective prior support, with
$\Pi(\mathfrak{M}_\Pi)=1$ and
$\Pi_n(\mathfrak{M}_\Pi)=1$ for $P_0^n$-almost every $D_n$. The
observational equivalence class of $P_0$ inside this support is
\begin{equation}
	\mathcal{E}_\Pi(P_0)
	=
	\{M\in\mathfrak{M}_\Pi : P_M=P_0\}.
	\label{eq:observational-equivalence-class}
\end{equation}
The feature profile is identifiable at $M_0$ under $\Pi$ if
$\tau(M)=\tau_0$ for all $M\in\mathcal{E}_\Pi(P_0)$. This is weaker than
identifying the full graph or the full SCM: observationally equivalent SCMs
may differ structurally while agreeing on the causal feature profile.

\paragraph{Identifiability assumptions (A1)--(A3).}
\begin{description}
	\item[(A1) Observational posterior consistency.]
	Let $P_\bullet:\mathfrak M\to\mathcal P(\mathcal Z)$ denote the map
	$M\mapsto P_M$. Then
	\begin{equation}
		(P_\bullet)_\#\Pi_n
		\Rightarrow
		\delta_{P_0}
		\quad\text{in }P_0^n\text{-probability}.
		\label{eq:observational-posterior-consistency}
	\end{equation}
where $\Rightarrow$ denotes weak convergence of probability laws on $\mathcal P(\mathcal Z)$.
  \item[(A2) Local continuity of the profile.]
For every $\epsilon>0$, there exists $\delta>0$ such that
\begin{equation}
	d_{\mathrm{BL}}(P_M,P_0)<\delta
	\quad\Longrightarrow\quad
	\|\tau(M)-\tau_0\|_2<\epsilon , \quad M\in\mathfrak{M}_\Pi.
	\label{eq:local-profile-continuity}
\end{equation}

	\item[(A3) Bounded profile.]
	There exists $B<\infty$ such that
	\begin{equation}
		\|\tau(M)\|_2\le B
		\quad\text{for all }M\in\mathfrak{M}_\Pi,
		\qquad
		\|\tau_0\|_2\le B .
		\label{eq:bounded-profile}
	\end{equation}
\end{description}

(A1) is a standard posterior consistency condition for
observational laws which typically holds in well-specified models under
Kullback--Leibler support and testing conditions. (A2) is the key identifiability assumption: continuity of the causal profile at $M_0$ when SCMs are compared through their observational laws. (A3) is a boundedness
condition used to pass from posterior concentration of $\tau(M)$ to convergence
of posterior means; it is imposed for simplicity and can be weakened to
posterior uniform integrability.

\paragraph{Equivalence-class disintegration.}
Assume the prior admits a regular conditional distribution over SCMs given their
observational law, restricted to the effective support $\mathfrak{M}_\Pi$, and
write
\[
\Pi_{P_0}(\cdot)=\Pi(\cdot\mid P_M=P_0).
\]
Let $Q_{\Pi,P_0}=\tau_\#\Pi_{P_0}$ be the induced distribution of causal
feature profiles among SCMs observationally equivalent to $P_0$.

\paragraph{(A4) Equivalence-class posterior convergence.}
For $D_n\sim P_0^n$,
\begin{equation}
	\tau_\#\Pi_n \Rightarrow Q_{\Pi,P_0}
	\quad\text{in }P_0^n\text{-probability},
	\label{eq:fiber-posterior-convergence}
\end{equation}
where $\Rightarrow$ denotes weak convergence of probability laws on $\mathbb{R}^p$.
This states that, after the observational law has been learned, the posterior over
causal profiles converges to the prior-induced distribution over causal profiles
inside $\mathcal{E}_\Pi(P_0)$.

\paragraph{Risk decomposition.}
For finite $n$, let $m_n(D_n)=\E_\Pi[\tau(M)\mid D_n]$ and
\[
V_n(D_n)
=
\E_\Pi\!\left[
\|\tau(M)-m_n(D_n)\|_2^2
\mid D_n
\right].
\]
By the squared-loss Pythagorean identity, every measurable estimator $f(D_n)$
satisfies
\begin{equation}
	R_2(f)
	=
	\E[V_n(D_n)]
	+
	\E\|f(D_n)-m_n(D_n)\|_2^2 .
	\label{eq:risk-decomposition}
\end{equation}
Thus $\E[V_n(D_n)]$ is the squared-loss risk that remains after predicting the
posterior mean exactly.

\subsection{Proof of Proposition~\ref{prop:bayes-action}}

\begin{proof}
	Condition on $D_n=d$ and write
	$m(d)=\E_{\Pi}[\tau(M)\mid D_n=d]$.  For any vector $z\in\mathbb{R}^p$,
	\begin{align}
		\E\!\left[\|z-\tau(M)\|_2^2\mid D_n=d\right]
		&=
		\|z-m(d)\|_2^2
		+
		\E\!\left[\|\tau(M)-m(d)\|_2^2\mid D_n=d\right].
	\end{align}
	The second term is independent of $z$, and the first term is uniquely
	minimized at $z=m(d)$.

	Fix a permutation $\pi$. By invariance of the prior and by the corresponding relabeling of samples, the joint law of the SCM and dataset is invariant under simultaneous feature relabeling:
	\[
	(M,D_n)\stackrel{d}{=}(M^\pi,D_n^\pi).
	\]
	Therefore conditioning on a relabeled dataset relabels the posterior over SCMs. Equivalently, for any integrable function $h$,
	\[
	\mathbb{E}_\Pi[h(M)\mid D_n=d^\pi]
	=
	\mathbb{E}_\Pi[h(M^\pi)\mid D_n=d]
	\]
	for almost every dataset $d$.
	
	Taking $h(M)=\tau_j(M)$ and using the equivariance of the causal feature profile,
	\[
	\tau_j(M^\pi)=\tau_{\pi(j)}(M),
	\]
	we obtain
	\[
	\begin{aligned}
		f_\Pi^\star(d^\pi)_j
		&=
		\mathbb{E}_\Pi[\tau_j(M)\mid D_n=d^\pi] \\
		&=
		\mathbb{E}_\Pi[\tau_j(M^\pi)\mid D_n=d] \\
		&=
		\mathbb{E}_\Pi[\tau_{\pi(j)}(M)\mid D_n=d] \\
		&=
		f_\Pi^\star(d)_{\pi(j)} .
	\end{aligned}
	\]
	Thus
	\[
	f_\Pi^\star(D_n^\pi)_j
	=
	f_\Pi^\star(D_n)_{\pi(j)},
	\qquad
	j=1,\ldots,p,
	\]
	almost surely.
\end{proof}

\subsection{Proof of Proposition~\ref{prop:profile-consistency}}
\begin{proof}
	Write
	\[
	\Pi_n(\cdot)=\Pi(\cdot\mid D_n),
	\qquad
	D_n\sim P_0^n .
	\]
	All convergence statements below are with respect to the sampling law $P_0^n$.

	Fix $\epsilon>0$. By \textup{(A2)}, there exists $\delta>0$ such that, for
	every $M\in\mathfrak{M}_\Pi$,
	\[
	d_{\mathrm{BL}}(P_M,P_0)<\delta
	\quad\Longrightarrow\quad
	\|\tau(M)-\tau_0\|_2<\epsilon .
	\]
	Thus, on $\mathfrak{M}_\Pi$,
	\[
	\{M:\|\tau(M)-\tau_0\|_2>\epsilon\}
	\subseteq
	\{M:d_{\mathrm{BL}}(P_M,P_0)\ge \delta\}.
	\]
	Since $\Pi_n(\mathfrak{M}_\Pi)=1$ for $P_0^n$-almost every $D_n$,
	\[
	\Pi_n\{M:\|\tau(M)-\tau_0\|_2>\epsilon\}
	\le
	\Pi_n\{M:d_{\mathrm{BL}}(P_M,P_0)\ge \delta\}
	\]
	for $P_0^n$-almost every $D_n$. The right-hand side converges to zero in
	$P_0^n$-probability by \textup{(A1)}, proving
	\cref{eq:profile-posterior-concentration}.
	
	Now assume also \textup{(A3)}. Let
	\[
	f_{\Pi,n}^\star(D_n)
	=
	\int \tau(M)\,d\Pi_n(M).
	\]
	For any $\eta>0$,
	\[
	\begin{aligned}
		\|f_{\Pi,n}^\star(D_n)-\tau_0\|_2
		&=
		\left\|
		\int \{\tau(M)-\tau_0\}\,d\Pi_n(M)
		\right\|_2  \\
		&\le
		\int \|\tau(M)-\tau_0\|_2\,d\Pi_n(M) \\
		&\le
		\eta
		+
		2B\,\Pi_n\{M:\|\tau(M)-\tau_0\|_2>\eta\}.
	\end{aligned}
	\]
	The final inequality uses \textup{(A3)} and
	$\Pi_n(\mathfrak{M}_\Pi)=1$: on $\mathfrak{M}_\Pi$,
	\[
	\|\tau(M)-\tau_0\|_2
	\le
	\|\tau(M)\|_2+\|\tau_0\|_2
	\le 2B .
	\]
	By \cref{eq:profile-posterior-concentration}, the posterior probability in
	the last display converges to zero in $P_0^n$-probability for every fixed
	$\eta>0$. 	Let $r>0$ and choose $\eta=r/2$. If $B=0$, then the claim is immediate.
	Otherwise,
	\[
	\begin{aligned}
		&P_0^n\!\left(
		\|f_{\Pi,n}^\star(D_n)-\tau_0\|_2>r
		\right) \\
		&\qquad\le
		P_0^n\!\left(
		\Pi_n\{M:\|\tau(M)-\tau_0\|_2>r/2\}
		> \frac{r}{4B}
		\right),
	\end{aligned}
	\]
	which converges to zero by \cref{eq:profile-posterior-concentration}. Hence
	\[
	f_{\Pi,n}^\star(D_n)
	\to
	\tau_0
	\quad\text{in }P_0^n\text{-probability}.
	\]
\end{proof}

\subsection{Empirical Illustration of \cref{prop:profile-consistency}}
\label{app:theory-nscale}
\begin{figure}[h]
  \centering
  \includegraphics[width=\linewidth]{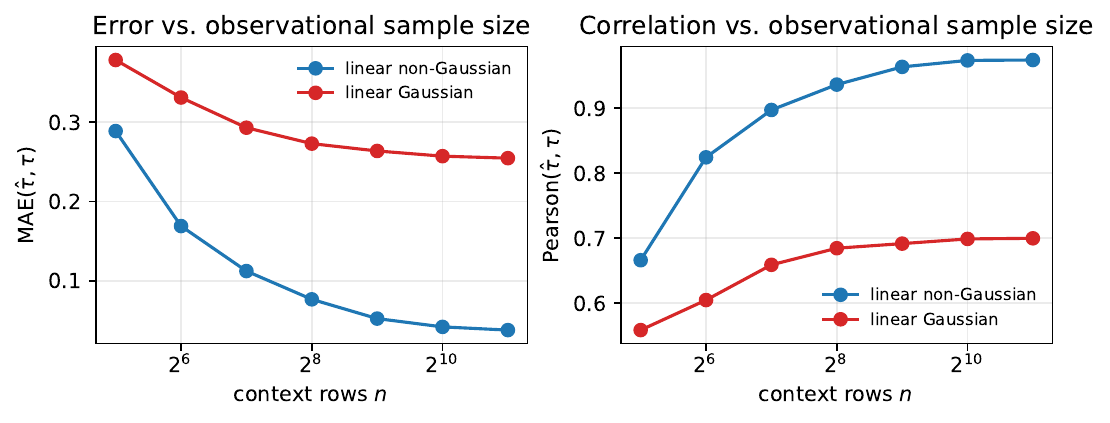}
  \caption{%
    Sample-size scaling of FADO under an identifiable prior
    (linear non-Gaussian, blue) and a non-identifiable prior (linear
    Gaussian, red).  Each point evaluates the prior-matched model
    on its own held-out prior across $150$ SCMs at $p{=}5$.  By
    \cref{prop:profile-consistency}
    $\hat\tau\to\tau$ as $n$ grows in the identifiable case; under non-identifiability, the Bayes
    action converges to the average of causal profiles in the equivalence class, which need not match $\tau(M_0)$
    for any individual SCM $M_0$ in that class.%
  }
  \label{fig:theory-nscale}
\end{figure}

Sample-size scaling of FADO under an identifiable prior (linear non-Gaussian, blue) and a non-identifiable prior (linear Gaussian, red), where the direction of edges cannot be determined from observational data, is given in Fig. \ref{fig:theory-nscale}.

\subsection{Non-identifiability and residual causal risk}
\label{app:nonidentifiability}

When the causal profile is not identifiable, infinite observational data reveal
$P_0$ but not which SCM in the equivalence class generated the data. The prior induces a distribution $Q_{\Pi,P_0}$ over causal feature profiles among SCMs observationally equivalent to $P_0$ (formal disintegration in \cref{app:assumptions}). Define $\bar{\tau}_\Pi(P_0)=\int t\,dQ_{\Pi,P_0}(t)$ and $U_\Pi(P_0)=\int \|t-\bar{\tau}_\Pi(P_0)\|_2^2\,dQ_{\Pi,P_0}(t)$. The causal profile is identifiable at $P_0$ under the prior exactly when $Q_{\Pi,P_0}$ is a point mass. Then $U_\Pi(P_0)=0$. Otherwise $U_\Pi(P_0)>0$, measuring residual variation in $\tau(M)$ among observationally equivalent SCMs.

For finite $n$, let $m_n(D_n)=\E_\Pi[\tau(M)\mid D_n]$ and $V_n(D_n)=\E_\Pi[\|\tau(M)-m_n(D_n)\|_2^2 \mid D_n]$. The squared-loss Pythagorean identity decomposes $\mathcal R_2(f)=\E[V_n(D_n)] + \E\|f(D_n)-m_n(D_n)\|_2^2$, so $\E[V_n(D_n)]$ is the irreducible portion of the risk after predicting the posterior mean exactly (\cref{eq:risk-decomposition}). We can now prove the following proposition which characterises the asymptotic behaviour of the Bayes action under non-identifiability -- we use an additional regularity assumption $\textup{(A4)}$, essentially requiring that the prior does not have a discontinuity in how it distributes causal profiles across nearby observational equivalence classes.

\begin{proposition}[Limit under non-identifiability]
	\label{prop:nonidentifiable-limit}
	Assume \textup{(A3)} and \textup{(A4)}. Then
	\begin{equation}
		m_n(D_n)\to \bar{\tau}_\Pi(P_0),
		\qquad
		V_n(D_n)\to U_\Pi(P_0),
		\label{eq:nonidentifiable-limit}
	\end{equation}
	both in $P_0^n$-probability. Moreover, for every observational estimator
	$g_n(D_n)$, the equivalence-class averaged risk
	\begin{equation}
		\mathcal{R}_{P_0}(g_n)
		=
		\int
		\E_{D_n\sim P_0^n}
		\|g_n(D_n)-t\|_2^2
		\,dQ_{\Pi,P_0}(t)
		\label{eq:fiber-averaged-risk}
	\end{equation}
	satisfies
	\begin{equation}
		\mathcal{R}_{P_0}(g_n)
		=
		U_\Pi(P_0)
		+
		\E_{D_n\sim P_0^n}
		\|g_n(D_n)-\bar{\tau}_\Pi(P_0)\|_2^2
		\ge
		U_\Pi(P_0).
		\label{eq:fiber-risk-lower-bound}
	\end{equation}
\end{proposition}

Thus, in non-identifiable regimes, the posterior mean converges to the
prior-weighted mean of the causal profiles compatible with $P_0$. This is the
squared-loss Bayes action under the pretraining prior, but it need not equal the
causal profile of any particular SCM in the equivalence class compatible with $P_0$.

\paragraph{Example: Two observationally equivalent SCMs.}
Suppose $M_1$ and $M_2$ satisfy $P_{M_1}=P_{M_2}=P_0$, but
$t_1=\tau(M_1)\neq t_2=\tau(M_2)$. For any estimator $g_n(D_n)$ based only on
observational samples,
\begin{equation}
	\max_{k\in\{1,2\}}
	\E_{D_n\sim P_0^n}
	\|g_n(D_n)-t_k\|_2^2
	\ge
	\frac{1}{4}\|t_1-t_2\|_2^2 .
	\label{eq:two-scm-lower-bound}
\end{equation}
If the prior assigns equal mass to $M_1$ and $M_2$, then
$\bar{\tau}_\Pi(P_0)=(t_1+t_2)/2$ and
$U_\Pi(P_0)=\frac{1}{4}\|t_1-t_2\|_2^2$, so the posterior mean attains
\cref{eq:two-scm-lower-bound} in the equivalence-class averaged sense.

\begin{figure}[t]
  \centering
  \includegraphics[width=\linewidth]{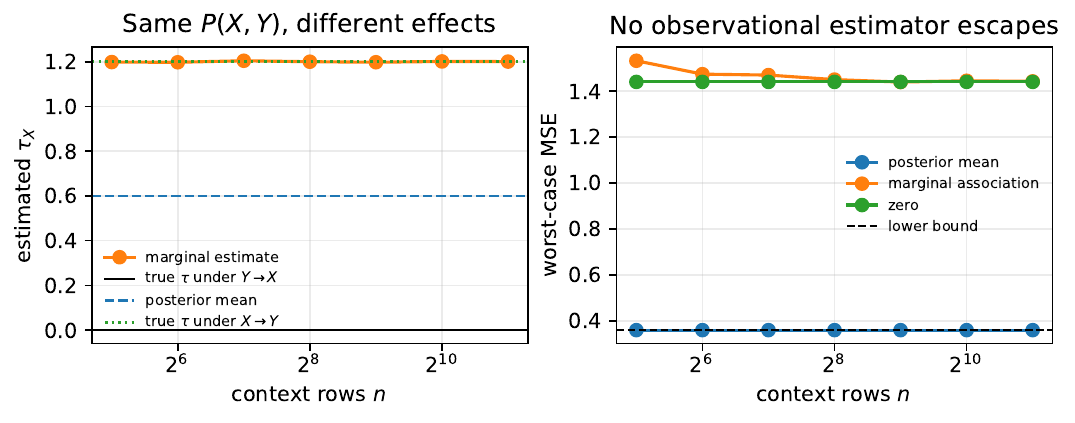}
  \caption{%
    Two Gaussian SCMs with the same observational covariance and different
    intervention effects ($M_1{:}~X\to Y$ with $\tau_X=2\rho$;
    $M_2{:}~Y\to X$ with $\tau_X=0$; $\rho=0.6$).  \emph{Left:} the marginal
    estimator targets $2\rho$ regardless of which SCM generated the data,
    matching $M_1$'s true effect and missing $M_2$'s; the equal-weight
    posterior mean is $\rho$ for both.  \emph{Right:} worst-case MSE across
    the two SCMs.  The posterior mean attains the
    \cref{prop:nonidentifiable-limit} lower bound $\rho^2{=}0.36$; marginal
    association and the zero estimator each lose to a different SCM.%
  }
  \label{fig:theory-toy-equivalence}
\end{figure}

The one-dimensional instance of this example is illustrated in Fig. \ref{fig:theory-toy-equivalence} and Table \ref{tab:theory-toy-equivalence}. Let $M_1$ be $X\to Y$ with standardized
effect $\tau_X=2\rho$, and let $M_2$ be $Y\to X$ with $\tau_X=0$. These SCMs
have the same observational covariance but different intervention effects. With
equal prior weights, the posterior mean is $\rho$, and the residual squared
error is $\rho^2$. For $\rho=0.6$, this gives $0.36$, matching the
no-free-lunch value in Table \ref{tab:theory-toy-equivalence}. Disagreement under hidden confounding, Gaussian direction ambiguity, or other
observational equivalences is therefore not merely a failure of amortization or
optimization. It is a limitation of the information available in $D_n$. A method
can resolve it only by adding identifying assumptions through the prior,
observing interventions, or abstaining.

\begin{table}[t]
  \centering
  \caption{%
    Per-SCM and worst-case MSE for the two-SCM Gaussian toy of
    \cref{fig:theory-toy-equivalence} at $n{=}2048$.  No observational
    estimator beats $\rho^2$ on the worst case
    (\cref{prop:nonidentifiable-limit}); the posterior mean meets the bound.%
  }
  \label{tab:theory-toy-equivalence}
  \input{figures/f_theory_toy_table.tex}
\end{table}

\subsection{Proof of Proposition~ \ref{prop:nonidentifiable-limit}}
\begin{proof}
	Let
	\[
	Q_n(D_n)=\tau_\#\Pi_n
	\]
	denote the posterior distribution of the causal feature profile. Then
	\[
	m_n(D_n)=\int t\,dQ_n(t),
	\qquad
	V_n(D_n)=\int \|t-m_n(D_n)\|_2^2\,dQ_n(t).
	\]
	
	We first record the squared-loss decomposition. For any measurable estimator
	$f(D_n)$, conditioning on $D_n$ gives
	\[
	\begin{aligned}
		\E\!\left[
		\|f(D_n)-\tau(M)\|_2^2
		\mid D_n
		\right]
		&=
		\E\!\left[
		\|f(D_n)-m_n(D_n)+m_n(D_n)-\tau(M)\|_2^2
		\mid D_n
		\right] \\
		&=
		\|f(D_n)-m_n(D_n)\|_2^2
		+
		\E\!\left[
		\|\tau(M)-m_n(D_n)\|_2^2
		\mid D_n
		\right],
	\end{aligned}
	\]
	because
	\[
	\E[\tau(M)-m_n(D_n)\mid D_n]=0.
	\]
	Taking expectations proves
	\[
	R_2(f)
	=
	\E[V_n(D_n)]
	+
	\E\|f(D_n)-m_n(D_n)\|_2^2 .
	\]
	
	We next prove the two convergence claims in \cref{eq:nonidentifiable-limit}.
	By \textup{(A3)}, both $Q_n(D_n)$ and $Q_{\Pi,P_0}$ are supported on the closed
	Euclidean ball $\{t:\|t\|_2\le B\}$ for $P_0^n$-almost every $D_n$. Weak convergence in \textup{(A4)} on a bounded support gives convergence of the first and second moments. Hence
	\[
	\int t_j\,dQ_n(t)
	\to
	\int t_j\,dQ_{\Pi,P_0}(t)
	\quad\text{in }P_0^n\text{-probability}
	\]
	for every $j$, and therefore
	\[
	m_n(D_n)=\int t\,dQ_n(t)
	\to
	\int t\,dQ_{\Pi,P_0}(t)
	=
	\bar{\tau}_\Pi(P_0)
	\quad\text{in }P_0^n\text{-probability}.
	\]
	
	Similarly, the map $t\mapsto \|t\|_2^2$ is bounded and Lipschitz on the ball
	$\{t:\|t\|_2\le B\}$. Thus
	\[
	\int \|t\|_2^2\,dQ_n(t)
	\to
	\int \|t\|_2^2\,dQ_{\Pi,P_0}(t)
	\quad\text{in }P_0^n\text{-probability}.
	\]
	Using
	\[
	V_n(D_n)
	=
	\int \|t\|_2^2\,dQ_n(t)
	-
	\left\|\int t\,dQ_n(t)\right\|_2^2
	\]
	and the convergence of the first moment just proved, we obtain
	\[
	V_n(D_n)
	\to
	\int \|t\|_2^2\,dQ_{\Pi,P_0}(t)
	-
	\left\|
	\int t\,dQ_{\Pi,P_0}(t)
	\right\|_2^2
	=
	U_\Pi(P_0)
	\]
	in $P_0^n$-probability.
	
	It remains to prove \cref{eq:fiber-risk-lower-bound}. For any fixed vector
	$a\in\mathbb{R}^p$,
	\[
	\begin{aligned}
		\int \|a-t\|_2^2\,dQ_{\Pi,P_0}(t)
		&=
		\int
		\|a-\bar{\tau}_\Pi(P_0)+\bar{\tau}_\Pi(P_0)-t\|_2^2
		\,dQ_{\Pi,P_0}(t) \\
		&=
		\|a-\bar{\tau}_\Pi(P_0)\|_2^2
		+
		\int
		\|t-\bar{\tau}_\Pi(P_0)\|_2^2
		\,dQ_{\Pi,P_0}(t),
	\end{aligned}
	\]
	because
	\[
	\int \{t-\bar{\tau}_\Pi(P_0)\}\,dQ_{\Pi,P_0}(t)=0.
	\]
	Substituting $a=g_n(D_n)$ and taking expectation over
	$D_n\sim P_0^n$ gives
	\[
	\mathcal{R}_{P_0}(g_n)
	=
	U_\Pi(P_0)
	+
	\E_{D_n\sim P_0^n}
	\|g_n(D_n)-\bar{\tau}_\Pi(P_0)\|_2^2
	\ge
	U_\Pi(P_0),
	\]
	as claimed.

For the Example with two observationally equivalent SCMs, since $P_{M_1}=P_{M_2}=P_0$, the distribution of
	$g_n(D_n)$ is the same whether the data are generated from $M_1$ or from
	$M_2$. Let $G$ denote a random vector with this common distribution, and write
	$t_k=\tau(M_k)$ for $k=1,2$. Then
	\[
	\begin{aligned}
		\E\|G-t_1\|_2^2+\E\|G-t_2\|_2^2
		&=
		2\E\left\|G-\frac{t_1+t_2}{2}\right\|_2^2
		+
		\frac{1}{2}\|t_1-t_2\|_2^2 .
	\end{aligned}
	\]
	Therefore,
	\[
	\begin{aligned}
		\max_{k\in\{1,2\}}
		\E\|G-t_k\|_2^2
		&\ge
		\frac{1}{2}
		\left(
		\E\|G-t_1\|_2^2+\E\|G-t_2\|_2^2
		\right) \\
		&=
		\E\left\|G-\frac{t_1+t_2}{2}\right\|_2^2
		+
		\frac{1}{4}\|t_1-t_2\|_2^2 \\
		&\ge
		\frac{1}{4}\|t_1-t_2\|_2^2 .
	\end{aligned}
	\]
	This proves \cref{eq:two-scm-lower-bound}.
	
	If the prior assigns equal mass to $M_1$ and $M_2$, then the induced
	equivalence-class distribution is
	\[
	Q_{\Pi,P_0}
	=
	\frac{1}{2}\delta_{t_1}
	+
	\frac{1}{2}\delta_{t_2}.
	\]
	Hence
	\[
	\bar{\tau}_\Pi(P_0)
	=
	\frac{t_1+t_2}{2},
	\]
	and
	\[
	\begin{aligned}
		U_\Pi(P_0)
		&=
		\frac{1}{2}
		\left\|
		t_1-\frac{t_1+t_2}{2}
		\right\|_2^2
		+
		\frac{1}{2}
		\left\|
		t_2-\frac{t_1+t_2}{2}
		\right\|_2^2  \\
		&=
		\frac{1}{4}\|t_1-t_2\|_2^2 .
	\end{aligned}
	\]
	Thus the posterior mean over the equivalence class attains the lower bound in
	the equivalence-class averaged sense.
\end{proof}

%% file: figures/f_theory_toy_table.tex
\begin{tabular}{lrrr}
\toprule
Estimator & MSE if $X\to Y$ & MSE if $Y\to X$ & Worst-case MSE \\
\midrule
Posterior mean & 0.360 & 0.360 & 0.360 \\
Marginal association & 0.001 & 1.442 & 1.442 \\
Zero effect & 1.440 & 0.000 & 1.440 \\
\midrule
No-free-lunch lower bound & & & 0.360 \\
\bottomrule
\end{tabular}